\documentclass{article}

\usepackage{arxiv}

\usepackage[utf8]{inputenc} 
\usepackage[T1]{fontenc}    
\usepackage{hyperref}       
\usepackage{url}            
\usepackage{booktabs}       
\usepackage{amsfonts}       
\usepackage{amsmath,amssymb}
\usepackage{nicefrac}       
\usepackage{microtype}      
\usepackage{lipsum}		
\usepackage{algorithm2e}
\usepackage{placeins}
\usepackage{graphicx}
\usepackage{natbib}
\usepackage{doi}
\usepackage{subcaption}
\usepackage{physics}
\usepackage{multirow}
\usepackage{colortbl}
\usepackage{supertabular}
\usepackage{xcolor}

\graphicspath{{imgs/}}

\title{Connectivity Optimized Nested Graph Networks for Crystal Structures}

\author{ Robin Ruff \\
	Institute of Theoretical Informatics \\
	Karlsruhe Institute of Technology \\
	Engler-Bunte-Ring 8, 
    76131 Karlsruhe, Germany \\
	\And
    Patrick Reiser\\
    Institute of Theoretical Informatics \\
    Karlsruhe Institute of Technology \\
    Engler-Bunte-Ring 8, 
    76131 Karlsruhe, Germany \\
    \And
	Jan Stühmer\\
    Institute for Anthropomatics and Robotics\\
    Karlsruhe Institute of Technology\\
    Engler-Bunte-Ring 8, 
    76131 Karlsruhe, Germany
    \vspace{0.2cm}\\
    Heidelberg Institute for Theoretical Studies\\
    Schloß-Wolfsbrunnenweg 35,\\
    69118 Heidelberg, Germany
    \And
    Pascal Friederich\\
    Institute of Theoretical Informatics \\
    Karlsruhe Institute of Technology \\
    Engler-Bunte-Ring 8,
    76131 Karlsruhe, Germany 
    \vspace{0.2cm} \\
    Institute of Nanotechnology \\
    Karlsruhe Institute of Technology \\
    Hermann-von-Helmholtz-Platz 1, \\
    76344 Eggenstein-Leopoldshafen, Germany \\
    \href{mailto:pascal.friederich@kit.edu}{\texttt{pascal.friederich@kit.edu}}
}

\renewcommand{\headeright}{}
\renewcommand{\undertitle}{}
\renewcommand{\shorttitle}{Connectivity Optimized Nested Graph Networks for Crystal Structures}

\hypersetup{
pdftitle={Connectivity Optimized Nested Graph Networks for Crystal Structures},
pdfsubject={GNN, Materials Science, cs.LG},
pdfauthor={Robin Ruff, Patrick Reiser, Jan Stühmer, Pascal Friederich},
pdfkeywords={Machine Learning, Graph Neural Networks, Materials Representations, Periodic Graphs, Nested Graph Neural Networks},
}

\begin{document}
\maketitle

\begin{abstract}
	Graph neural networks (GNNs) have been applied to a large variety of applications in materials science and chemistry. Here, we recapitulate the graph construction for crystalline (periodic) materials and investigate its impact on the GNNs model performance. We suggest the asymmetric unit cell as a representation to reduce the number of atoms by using all symmetries of the system. This substantially reduced the computational cost and thus time needed to train large graph neural networks without any loss in accuracy. Furthermore, with a simple but systematically built GNN architecture based on message passing and line graph templates, we introduce a general architecture (Nested Graph Network, NGN) that is applicable to a wide range of tasks. We show that our suggested models systematically improve state-of-the-art results across all tasks within the MatBench benchmark. Further analysis shows that optimized connectivity and deeper message functions are responsible for the improvement. Asymmetric unit cells and connectivity optimization can be generally applied to (crystal) graph networks, while our suggested nested graph framework will open new ways of systematic comparison of GNN architectures.
\end{abstract}

\keywords{Machine Learning, Graph Neural Networks, Materials Representations, Periodic Graphs, Nested Graph Neural Networks}

\section{Introduction}

Since seminal work by \citet{duvenaud2015convolutional}, graph neural networks (GNNs) developed into one of the most versatile and accurate classes of machine learning models for the prediction of molecular and material properties. 
Consequently, GNNs find increasing application in materials sciences for structure-property predictions~\citep{PhysRevMaterials.4.093801}, materials screening~\citep{SchmidtCrystalAttention2021} and high-throughput simulations~\citep{Behler2011NP}. Learning on experimental or simulated databases~\citep{kirklin_open_2015,ocp_dataset,oc22_dataset, choudhary_joint_2020}, GNNs show promising potential to develop new materials to tackle societies growing demand for high-performance materials in the fields of catalysis, renewable energies, energy conversion or functional materials~\citep{gielen_climate_2016, doi:10.1021/acs.accounts.5b00530, doi:10.1021/acs.chemmater.9b02166}. 

Graph convolutional neural networks operate on the (spatial) graph structure to transform node embeddings and have been suggested for semi-supervised node classification~\citep{kipf-welling-gnn, kipf_gcn_relation_2018}.
With the focus on molecular graphs, the message passing framework (MPNN) was proposed by~\citet{gilmer2017neural} in order to group and generalize many GNN architectures that update node representations by iteratively exchanging information, or messages, across edges in the graph~\citep{velickovic2018graph, GraphSAGE}.
Neural networks designed to predict the potential energy surface of molecules for molecular dynamics simulations~\citep{NNPBehlerForce}, such as the continuous-filter convolutional network SchNet~\citep{schutt2017schnet}, can also be interpreted as MPNN graph networks.
Significant progress was achieved by incorporating further geometric information such as bond~\citep{Klicpera2020DimeNet} and dihedral angles~\citep{klicpera2021gemnet} into the MPNN scheme. To compute explicit angle information, the graph network has to operate on higher order pathways or connect edge information in a so-called line graph, which sets up a graph on edges $L(G)$, i.e.\ on-top of the underlying graph $G$~\citep{chen2018supervised}. A state-of-the-art model which makes use of this principle and reaches good performance for materials is ALIGNN~\citep{choudhary2021atomistic}.
Recently, equivariant graph networks have been introduced~\citep{thomas2018TFN}, which build on irreducible representations of the euclidean symmetry groups to achieve equivariance under e.g. rotation and translation~\citep{batzner20223}.
Although many of the previously mentioned GNNs can or have been applied to materials, fewer architectures are developed with a primary focus on crystalline systems. 
The crystal-graph convolution neural network (CGCNN) first introduced a GNN architecture on a crystalline system by constructing a multi-graph that correctly represents the atomic neighbors in a periodic system~\citep{xie2018crystal}. Its improved version iCGCNN incorporates information on the Voronoi tessellated crystal structure and explicit three-body correlations of neighboring constituent atoms~\citep{park2020developing}. MEGNet further leverages global state information and added edge updates in the convolution process~\citep{chen2019graph}. With GeoCGNN a geometric GNN was introduced~\citep{chengGeoCGNN2021}, which encodes the local geometrical information using an attention mask composed of Gaussian radial basis functions and plane waves from a k-point mesh of Monkhorst Pack special points~\citep{PhysRevB.13.5188}. 
Although many GNN model architectures have been proposed in this context~\citep{xie2018crystal, chen2019graph, yamamoto2019crystal, park2020developing, chengGeoCGNN2021, choudhary2021atomistic, chen2022universal}, there does not yet seem to be unanimous consent in the literature on which is the best method or the most decisive tool in geometric deep learning to process crystalline materials.

In most cases, newly introduced GNNs make specific improvements over previous architectures and propose multiple reasonable new design decisions inspired by chemical domain knowledge.
While this approach has so far led to a consistent improvement of model accuracy, we choose a more systematic approach inspired by the work of~\citet{you2020design}:
First, we stake out a new design space based on an extension of the original graph network (GN) framework~\citep{battaglia2018relational}, namely nested graph networks (NGNs).
Then we navigate through the design space to find suitable NGN architectures.

In this work, we re-evaluate edge selection methods to build a multi-graph as suggested by~\citet{xie2018crystal} and compare their performance on a wide set of parameters. In this context, we introduce the \textit{asymmetric unit cell} as a representation to further exploit crystal symmetries and effectively reduce the number of edges. Next, we develop a connectivity-optimized crystal graph network (coGN) from message passing and line-graph templates which are intentionally kept as general as possible. By optimizing within the generalized  family of GNNs, we improve state-of-the-art results on 6 out of 9 of the MatBench benchmark datasets~\citep{dunn_benchmarking_2020} and achieve close to parity results with the best models on the remaining datasets, making our model the best general model on the entire set of diverse tasks in the benchmark.

\section{Crystal Graph Construction}

There are two main challenges when trying to build graph representations of crystal structures in contrast to organic molecules:
(a) Bonds between atoms in crystals have more diverse types (covalent, ionic, metallic bonds), or are often not well defined at all.
(b) Crystal structures have no fixed finite size, as they are defined as periodic repetitions of a unit cell.

\paragraph{Edge selection \label{sec:edge_selection}}
The first aspect raises the question of which edges to add to the graph that describes the crystal.
To circumvent this problem one relies on the geometrical properties of the atom arrangement instead of chemically informed bonds.

\begin{figure}[h]
	\centering
	\begin{subfigure}[b]{0.25\textwidth}
		\centering
		\def\svgwidth{0.7\textwidth}
		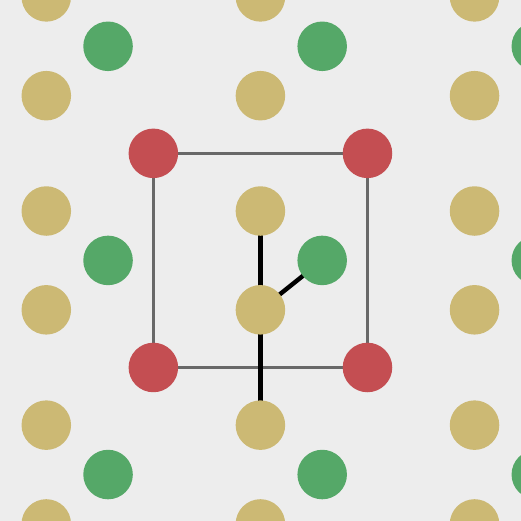
		\caption{$k$NN Edges}
		\label{fig:knn-edges}
	\end{subfigure}
	\begin{subfigure}[b]{0.25\textwidth}
		\centering
		\def\svgwidth{0.7\textwidth}
		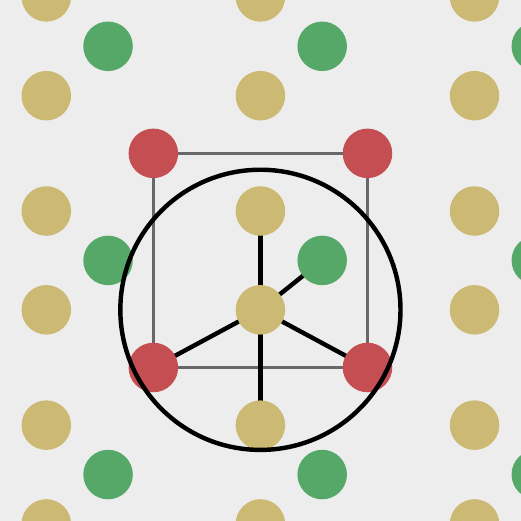
		\caption{Radius Edges}
		\label{fig:radius-edges}
	\end{subfigure}
	\begin{subfigure}[b]{0.25\textwidth}
		\centering
		\def\svgwidth{0.7\textwidth}
		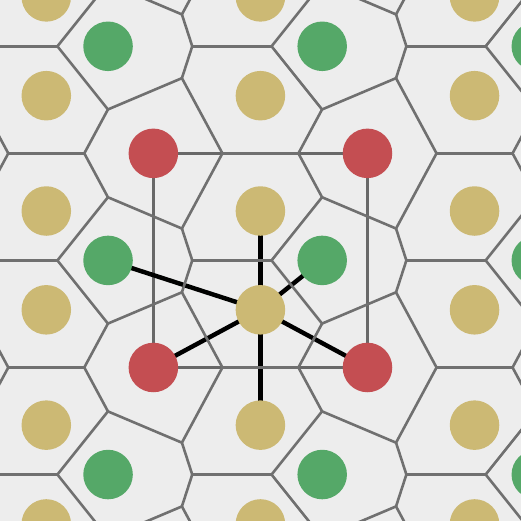
		\caption{Voronoi Edges}
		\label{fig:voronoi-edges}
	\end{subfigure}
	\caption{Edge selection for a single atom in a crystal structure. Red atoms mark the cubic unit cell.}
	\label{fig:edge-selection}
\end{figure}

Figure~\ref{fig:edge-selection} schematically shows a two-dimensional crystal pattern and three different methods for selecting edges between atoms.
The k-nearest-neighbors approach (Figure~\ref{fig:knn-edges}) depends on the number of neighbors $k$, but can lead to largely different edge distances when the crystal density varies.
The radius-based approach (Figure~\ref{fig:radius-edges}) limits the distance between two nodes by a hyperparameter $r$,
but the number of neighbors is unbounded and the method can lead to either disconnected or overly dense graphs if $r$ is chosen inappropriately.
The parameter-free Voronoi-based approach (Figure~\ref{fig:voronoi-edges}) leads to an intuitive edge selection where edges are drawn between two atoms if there is a Voronoi cell ridge between them.  However, at least in theory, the number of edges and their distances are also unbounded for this approach.

All three edge selection methods have been used previously in the context of crystals and GNNs \citep{chen2019graph, park2020developing, xie2018crystal, isayev2015materials}.
But to our knowledge, there is no detailed comparison between the methods and hyperparameters.

\begin{figure}[h]
	\centering
	\begin{subfigure}[b]{0.25\textwidth}
		\centering
		\def\svgwidth{0.7\textwidth}
		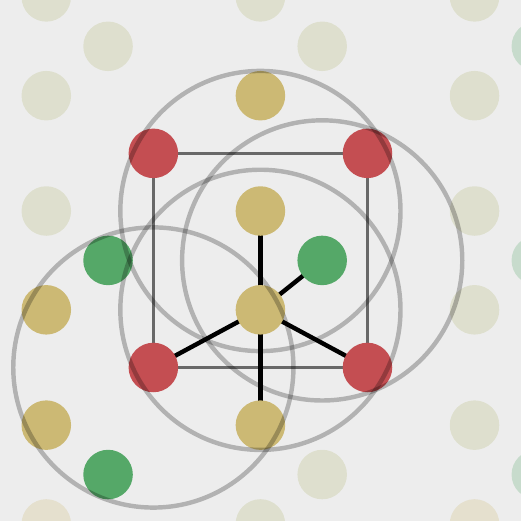
		\caption{No symmetries}
		\label{fig:np}
	\end{subfigure}
	\begin{subfigure}[b]{0.25\textwidth}
		\centering
		\def\svgwidth{0.7\textwidth}
		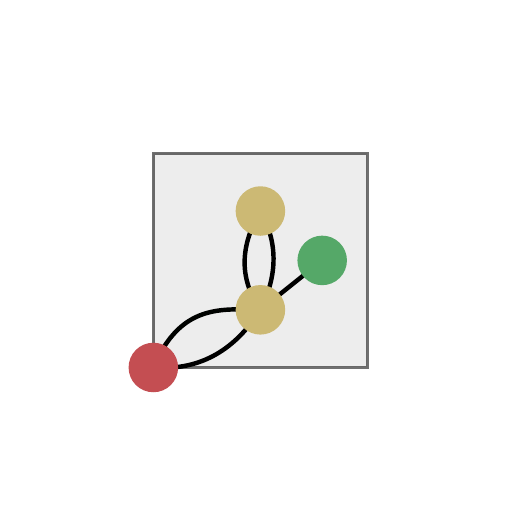
		\caption{Periodicity}
		\label{fig:unit}
	\end{subfigure}
	\begin{subfigure}[b]{0.25\textwidth}
		\centering
		\def\svgwidth{0.7\textwidth}
		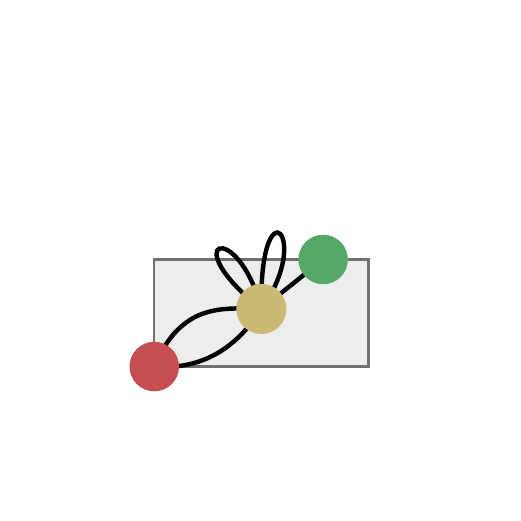
		\caption{All symmetries}
		\label{fig:asu}
	\end{subfigure}
	\caption{Graph representations of crystal structures exploiting different levels of symmetries. The \textit{asymmetric unit cell} is shown in (c)}
	\label{fig:symmetries}
\end{figure}

\paragraph{Exploiting crystal symmetries}
\label{sec:crystal-symmetries}
In contrast to molecules, crystal structures are modeled as (infinite) periodic repetitions of the unit cell atom motif.
A direct approach to extracting a finite graph for a given crystal structure is to simply select all unit cell atoms with their respective neighbors (Figure~\ref{fig:np}).
In an information-theoretical sense, this representation captures the entire information of the crystal for a given edge selection method.
However, from a naive message-passing perspective, only the nodes representing atoms inside of the central unit cell receive messages from all their neighboring atom nodes, which would model a finite graph rather than an infinite periodic graph.

To solve this issue, \citet{xie2018crystal} proposed a multigraph crystal representation (Figure~\ref{fig:unit}) introducing periodic/cyclic boundary conditions, which we will refer to as a unit cell graph.
In this representation, one node represents all the shift-equivalent atoms of the crystal, which usually results in multiple edges with distance information matching their translated lattice positions.
As a consequence, GNNs will always provide equivalent node embedding for periodic atoms, consistent with Bloch's theorem.

The periodicity of the unit cell graph represents translation symmetry.
Since crystals often exhibit more symmetries, we propose the asymmetric unit graph representation for crystals, which considers all symmetries of a crystal structure.\footnote{The set of symmetries for crystals and thus the asymmetric unit cell representation can be determined automatically based on the normal unit cell~\citep{togo2018texttt}.}
In this representation, all symmetry-equivalent atoms are represented by a single node in the graph.
The example crystal in Figure~\ref{fig:symmetries} exhibits a horizontal reflection symmetry.
The two yellow atoms in the crystal unit cell are symmetry-equivalent and therefore merged into one node (with multiplicity two) for the asymmetric unit cell (Figure~\ref{fig:asu}).
Just like in unit cell graphs, self-loops and multi-edges can occur in asymmetric unit cell graphs.

%

Since physical target properties in ML tasks are often invariant under E(3) symmetry operations, many GNNs are designed to be E(3)-invariant, but as consequence, yield equal node embeddings for symmetrical atoms in the unit cell graph, leading to redundant computations in the message passing step.
The asymmetric unit graph representation can further remove these redundancies and yet maintain the same node embeddings $x_v$.\footnote{It is in principle also possible to adapt E(3)-equivariant GNNs to asymmetric unit graph representations, by specifying equivariant convolutional layers on the asymmetric unit graph and adapting message passing accordingly.}
However, global readout operations have to be adapted to handle asymmetric unit cells, since atoms can have different symmetry-related multiplicities $\mathfrak{m}_v$ (the number of symmetry equivalent atoms for each equivalence class)
A simple adaptation of the readout function can fix the issue and restore equal results for unit cell graphs and asymmetric unit graphs:
\begin{equation*}\hypertarget{eq:asu-aggregation}{}{
		\underset{v \in V}{\text{agg}'(x_v)} = \begin{cases}
			\underset{v \in V}{\text{agg}}(x_v \cdot \mathfrak{m}_v) \cdot \frac{|V|}{\sum_v \mathfrak{m}_v} & \text{for mean or attention} \\
			\underset{v \in V}{\text{agg}}(x_v \cdot \mathfrak{m}_v) & \text{for sum} \\
			\underset{v \in V}{\text{agg}}(x_v) & \text{for min or max} \\
\end{cases}
}\label{eq:asu-aggregation}
\end{equation*}

\section{Nested Graph Network Framework}

In addition to the choice of input representation, the GNN model architecture has a substantial impact on the quality of crystal graph property predictions.

To find the best GNN architecture for a certain task, it is instructive to systematically explore (see e.g. \citep{you2020design}) the design space of GNN modules and building blocks. Moreover, a framework has to be chosen on how to define and process GNN modules. The Message Passing Framework by \citet{gilmer2017neural}, for example, has shown that a framework can unify and accelerate the efforts of the scientific community. 


For a combinatorial generalization of GNNs, we build upon the graph network (GN) framework of \citet{battaglia2018relational} in this work.
In this framework, one GNN layer is described by a GN block, which transforms a generic attributed graph with edge, node, and global graph features via three update functions $\phi$ and three aggregation functions $\rho$.

However, many state-of-the-art models such as DimeNet~\citep{klicpera2020directional, klicpera2020directional}, ALIGNN~\citep{choudhary2021atomistic} and M3GNet~\citep{chen2022universal}, which incorporate many-body interactions between atoms, are only indirectly captured by GNs.
Therefore we propose an extension to the framework which we term nested graph networks (NGNs). 
In the following, we introduce NGNs by first discussing how angle information is incorporated via the line graph concept~\citep{choudhary2021atomistic} and then explaining the flow of NGN calculations, before we will show concrete examples of implementations of NGNs in Section~\ref{sec:search}.

\begin{figure}[h]
	\centering
	\def\svgwidth{0.55\textwidth}
	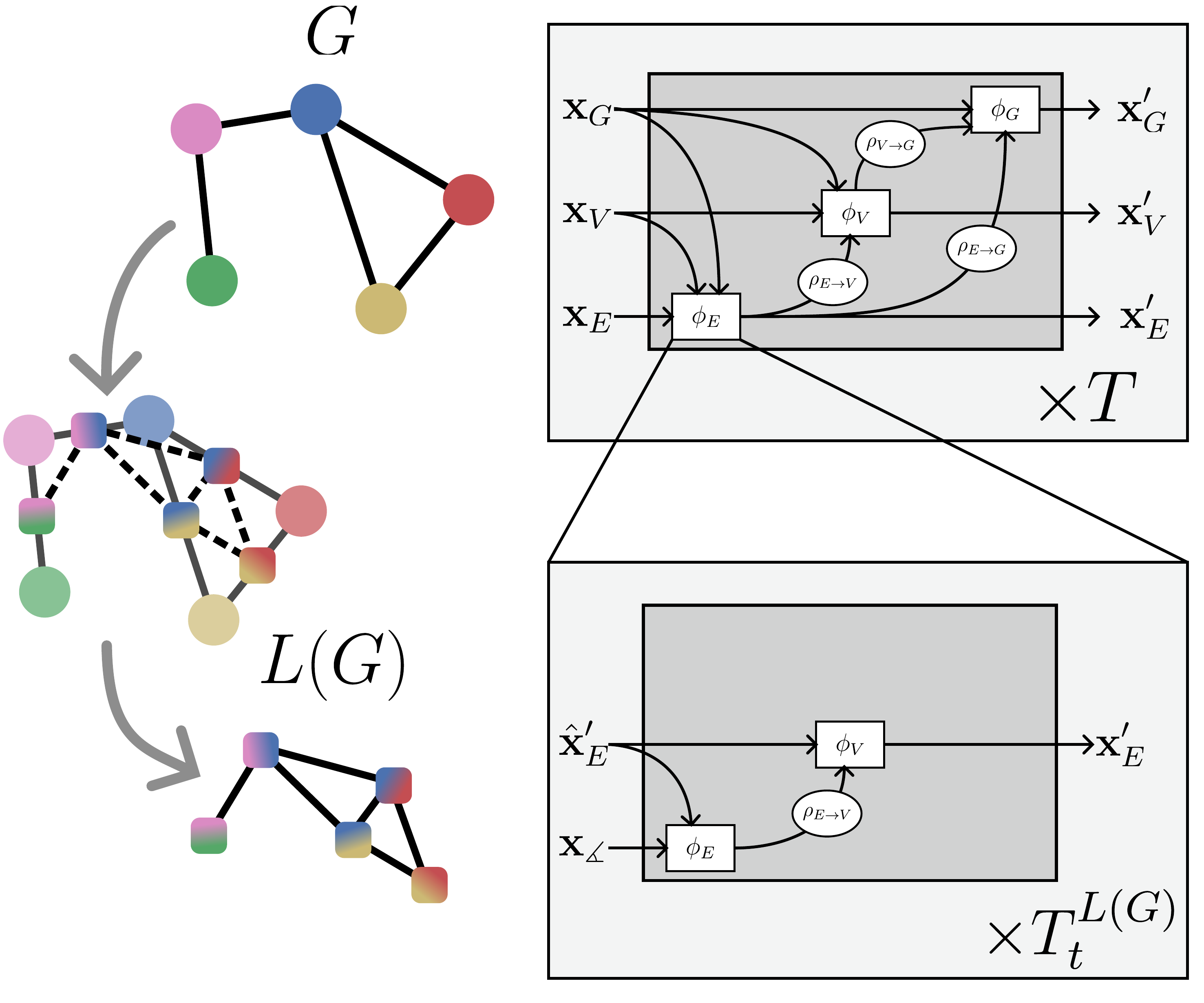
	\caption{On the left: A graph $G$ and the construction of its line graph $L(G)$. On the right: Nested graph network architecture with GN blocks working on $G$ and containing other GN blocks working on the line graph $L(G)$ as edge update function $\phi_E$. The line graph is able to process multi-node geometric features, such as angles (see Table~\ref{tab:graph-entities})}
	\label{fig:nestedgn}
\end{figure}

To achieve rotation invariance, popular models such as SchNet~\citep{schutt2017schnet} or MEGNet~\citep{chen2019graph} only include scalar distances between two atoms as edge features.
However, to incorporate more geometric information, models such as DimeNet \citep{klicpera2020directional} or ALIGNN~\citep{choudhary2021atomistic} additionally use E(3)-invariant angle information, represented by combinations of edges (see Figure~\ref{fig:expressiveness}).
In the line graph $L(G)$, which can be constructed uniquely based on G (see Figure~\ref{fig:nestedgn} and \citet{harary1960some}), there is an edge $e^{L(G)}_{e_{ij}, e_{jk}}$ for every two incident edges $e_{ij}, e_{jk}$ in $G$.
This enables the assignment of angular information between three atoms to the edges of the line graph. The same applies to (generalized) dihedral angles (4-node or 3-edge objects) in the second-order line graph $L(L(G))$.

\begin{table}[!ht]
    \centering
    \caption{Correspondence between entities in the crystal, the Crystal Graph $G$, its Line Graph $L(G)$, etc.}
    \begin{tabular}{llll}
		\toprule
        Entity in Crystal & $G$ & $L(G)$ & $L(L(G))$\\ 
		\midrule
        Atoms   & Nodes & ~ & ~\\ 
        Bonds & Edges & Nodes & ~\\ 
        Angles & ~ & Edges & Nodes\\
        Dihedrals & ~ & ~ & Edges \\ 
		\bottomrule
    \end{tabular}
	\label{tab:graph-entities}
\end{table}

\begin{figure}[h]
\hypertarget{fig:generalized-models}{%
\centering
		\def\svgwidth{\textwidth}
\begingroup%
  \makeatletter%
  \providecommand\color[2][]{%
    \errmessage{(Inkscape) Color is used for the text in Inkscape, but the package 'color.sty' is not loaded}%
    \renewcommand\color[2][]{}%
  }%
  \providecommand\transparent[1]{%
    \errmessage{(Inkscape) Transparency is used (non-zero) for the text in Inkscape, but the package 'transparent.sty' is not loaded}%
    \renewcommand\transparent[1]{}%
  }%
  \providecommand\rotatebox[2]{#2}%
  \newcommand*\fsize{\dimexpr\f@size pt\relax}%
  \newcommand*\lineheight[1]{\fontsize{\fsize}{#1\fsize}\selectfont}%
  \ifx\svgwidth\undefined%
    \setlength{\unitlength}{3089.76377953bp}%
    \ifx\svgscale\undefined%
      \relax%
    \else%
      \setlength{\unitlength}{\unitlength * \real{\svgscale}}%
    \fi%
  \else%
    \setlength{\unitlength}{\svgwidth}%
  \fi%
  \global\let\svgwidth\undefined%
  \global\let\svgscale\undefined%
  \makeatother%
  \begin{picture}(1,0.26605505)%
    \lineheight{1}%
    \setlength\tabcolsep{0pt}%
    \put(0,0){\includegraphics[width=\unitlength,page=1]{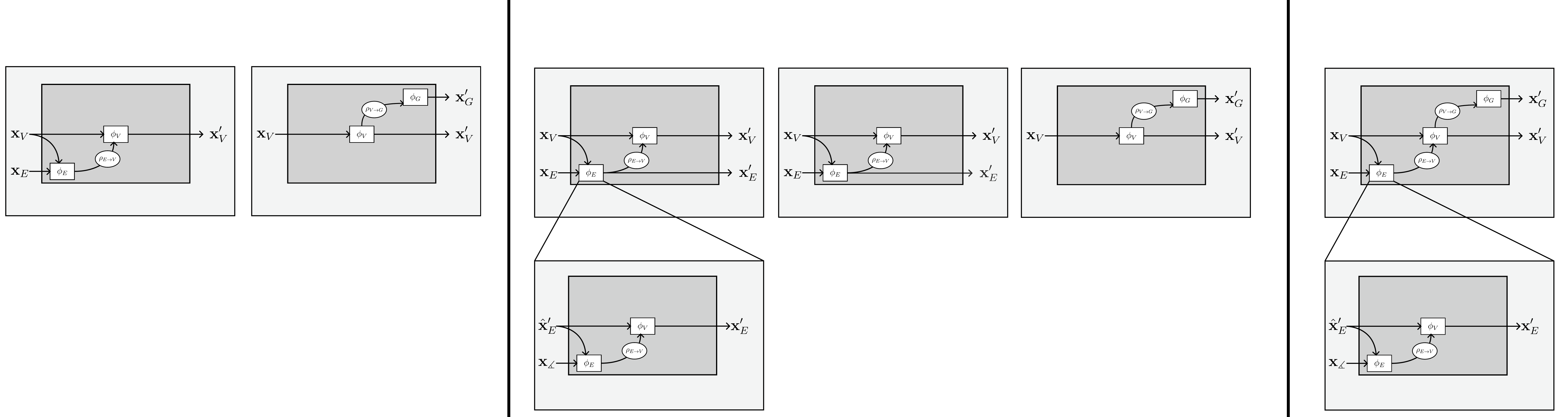}}%
    \put(0.15639399,0.23931426){\color[rgb]{0,0,0}\makebox(0,0)[t]{\lineheight{1.25}\smash{\begin{tabular}[t]{c}SchNet\end{tabular}}}}%
    \put(0.56724215,0.23931426){\color[rgb]{0,0,0}\makebox(0,0)[t]{\lineheight{1.25}\smash{\begin{tabular}[t]{c}ALIGNN\end{tabular}}}}%
    \put(0.91422685,0.23931426){\color[rgb]{0,0,0}\makebox(0,0)[t]{\lineheight{1.25}\smash{\begin{tabular}[t]{c}DimeNet\end{tabular}}}}%
    \put(0,0){\includegraphics[width=\unitlength,page=2]{nested_gn_models.pdf}}%
  \end{picture}%
\endgroup%

\caption{Other models in the NGN framework.}\label{fig:generalized-models}
}
\end{figure}

NGNs operate on the graph $G$ as well as the line graph $L(G)$ (potentially also $L(L(G))$ etc.), exploiting the one-to-one mapping of edges in $G$ and nodes in $L(G)$ (see Table~\ref{tab:graph-entities}).
Each edge update function $\phi_E$ in GN blocks that operate on $G$ can be instantiated as a nested GN (see Figure~\ref{fig:nestedgn}).
A more detailed description of the algorithm can be found in Appendix~\ref{app:ngnalg}.
The NGN framework extends the usual sequential compositionality of GN blocks with a hierarchical/nested compositionality, thereby increasing the expressiveness \citep{maron2019provably}. 
The composition of simple and well-known building blocks facilitates the implementation and the ease of understanding of the framework.
Furthermore, NGNs generalize existing models such as SchNet \citep{schutt2017schnet}, DimeNet \citep{klicpera2020directional} and ALIGNN \citep{choudhary2021atomistic}, making them more comparable and extensible (see Figure \ref{fig:generalized-models}). 

\section{Line Graph Construction} \label{sec:linegraph}

\begin{figure}[h]
	\centering
	\begin{subfigure}[b]{0.30\textwidth}
		\centering
		\def\svgwidth{0.8\textwidth}
		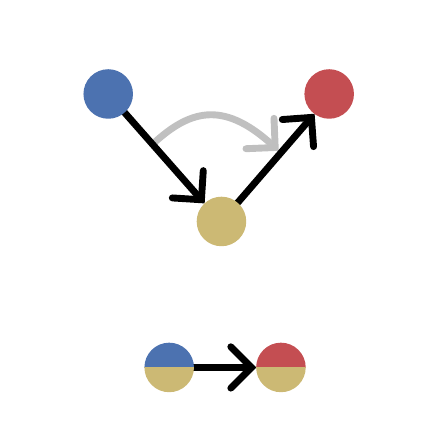
		\caption{Original line graph.}
		\label{fig:linegraph-variant1}
	\end{subfigure}
	\begin{subfigure}[b]{0.30\textwidth}
		\centering
		\def\svgwidth{0.8\textwidth}
		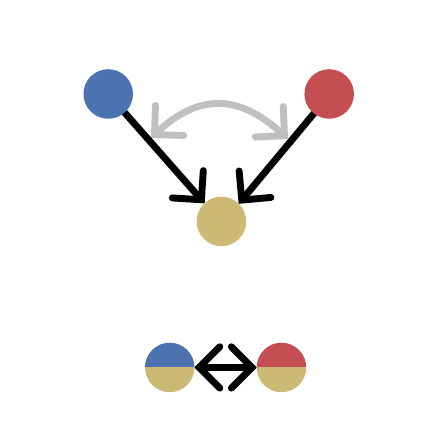
		\caption{Our line graph variant.}
		\label{fig:linegraph-variant2}
	\end{subfigure}
	\caption{Line graph variants.}
	\label{fig:linegraph-variants}
\end{figure}

Mathematically a line graph for a directed graph is defined as follows:
An edge exists in \(L(G)\) for each corresponding edge pair \((e_{ij}, e_{jk})\), which forms a path of length two
in \(G\) (Figure~\ref{fig:linegraph-variant1}).
This definition coincides with the angles \(\measuredangle e_{ij}e_{jk}\) as used in in DimeNet
\citep{klicpera2020directional}, GemNet
\citep{gasteiger2021gemnet} and ALIGNN
\citep{choudhary2021atomistic}.
In this work, we propose a deviation from the original line graph definition and use angles \(\measuredangle e_{ij}e_{kj}\) between edges that have the same destination node, instead of edges that form a path of length two (see Figure~\ref{fig:linegraph-variant2}).
The intuition behind this deviation is based on the interpretation of edge messages in the GNN as force contributions that act on an atom from different directions~\citep{Defferrard2020DeepSphere}.

Note that we will use the term line graph very loosely in this work.
Even though the variant is strictly speaking not a line graph we will still refer to it as such.
In general, we will use the term line graph for every graph $L(G)$ that has a bijection between edges in $G$ and nodes in $L(G)$ and a deterministic method for constructing edges in $L(G)$ based on the topology of $G$.

This generalized definition would also allow for incorporating dihedral angles by adding edges $\measuredangle e_{ij}e_{kl}$ to the line graph for every path $(e_{ij}, e_{jk}, e_{kl})$ of length three in the graph $G$ \citep{hsu2021efficient, klicpera2020fast, klicpera2021gemnet}.
Another way to incorporate information of dihedral edges into the model would be to construct the second-order line graph $L(L(G))$ and add another layer of nesting to the model architecture. 

\section{Evaluation}\label{sec:evaluation}

\paragraph{Datasets}{To evaluate the preprocessing methods and GN architectures we relied on the MatBench benchmark~\citep{dunn_benchmarking_2020}, which can be considered to be the ImageNet~\citep{ILSVRC15} for machine learning models in materials science. The benchmark currently consists of 13 strictly standardized supervised crystal property prediction tasks curated from different data
sources~\citep{jainn2013Genome, de_jong_charting_2015, choudhary_high_throughput_2017, C2EE22341D}.
Out of the 13 tasks, four provide only the crystal composition as input
and the other nine incorporate the crystal structure with the geometric
arrangement of atoms. Since this work focuses on crystal structures
specifically, we only use the nine datasets that rely on structural
information for property prediction.
The size of the datasets ranges from 636 to 132,752 crystal instances.}

\paragraph{Implementations}{

We used the Keras Graph Convolution Neural Networks (KGCNN) library~\citep{reiser2021graph} to implement the GN and NGNs used in the experiments of this work.
The code for crystal preprocessing and GNN models is available online\footnote{\url{https://github.com/matbench-submission-coGN/CrystalGNNs}}
}

\subsection{GNN Architecture Search}\label{sec:search}

To find a suitable GN architecture, we conduct an architecture search within the NGN framework.
In order to generalize previous models, our NGN framework is very flexible and spans a large hyperparameter space for architectural decisions but which makes the exploration of the entire space infeasible. If no nesting is explicitly required, the NGN framework falls back to the GN framework of ~\citet{battaglia2018relational}. For comparison, we search architectures with and without nesting, which lead to the connectivity optimized graph architectures \textbf{coGN} and \textbf{coNGN}, respectively. The search procedure and architecture details is discussed in the following.

\paragraph{Graph Network Exploration}{

We start with a simple non-nested architecture shown in Figure~\ref{fig:archgn},
which consists of sequentially connected GN blocks, which can be divided into three phases.

\begin{figure*}[h]
	\centering
	\def\svgwidth{.9\textwidth}
\begingroup%
  \makeatletter%
  \providecommand\color[2][]{%
    \errmessage{(Inkscape) Color is used for the text in Inkscape, but the package 'color.sty' is not loaded}%
    \renewcommand\color[2][]{}%
  }%
  \providecommand\transparent[1]{%
    \errmessage{(Inkscape) Transparency is used (non-zero) for the text in Inkscape, but the package 'transparent.sty' is not loaded}%
    \renewcommand\transparent[1]{}%
  }%
  \providecommand\rotatebox[2]{#2}%
  \newcommand*\fsize{\dimexpr\f@size pt\relax}%
  \newcommand*\lineheight[1]{\fontsize{\fsize}{#1\fsize}\selectfont}%
  \ifx\svgwidth\undefined%
    \setlength{\unitlength}{1502.36220472bp}%
    \ifx\svgscale\undefined%
      \relax%
    \else%
      \setlength{\unitlength}{\unitlength * \real{\svgscale}}%
    \fi%
  \else%
    \setlength{\unitlength}{\svgwidth}%
  \fi%
  \global\let\svgwidth\undefined%
  \global\let\svgscale\undefined%
  \makeatother%
  \begin{picture}(1,0.22641509)%
    \lineheight{1}%
    \setlength\tabcolsep{0pt}%
    \put(0,0){\includegraphics[width=\unitlength,page=1]{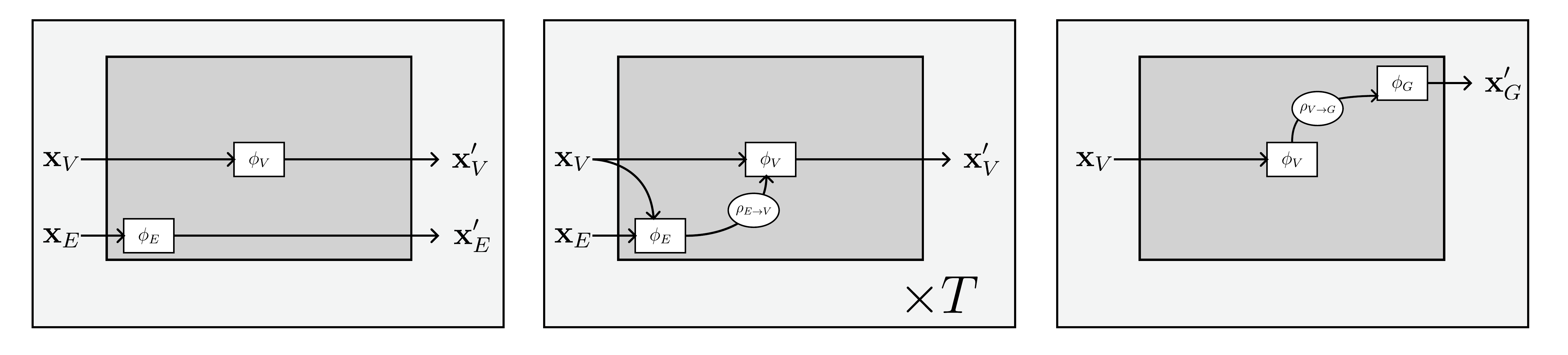}}%
    \put(0.02576297,0.02474771){\color[rgb]{0,0,0}\makebox(0,0)[lt]{\lineheight{1.25}\smash{\begin{tabular}[t]{l}Embedding Block\end{tabular}}}}%
    \put(0.35144935,0.02474771){\color[rgb]{0,0,0}\makebox(0,0)[lt]{\lineheight{1.25}\smash{\begin{tabular}[t]{l}Processing Blocks\end{tabular}}}}%
    \put(0.67905498,0.0237285){\color[rgb]{0,0,0}\makebox(0,0)[lt]{\lineheight{1.25}\smash{\begin{tabular}[t]{l}Readout Block\end{tabular}}}}%
    \put(0,0){\includegraphics[width=\unitlength,page=2]{gn_basic_structure.pdf}}%
  \end{picture}%
\endgroup%

	\caption{Basic graph network structure.}
	\label{fig:archgn}
\end{figure*}

Moreover, we optimize the hyperparameter space on one dataset only, namely the \texttt{log\_gvrh} MatBench dataset, and tested the model on all other datasets of MatBench without changing hyperparameters to verify that the found architecture is indeed a suitable candidate for multiple diverse tasks on crystal graphs.

The first block embeds atom/node features and bond/edge features independent from one another.
In the second phase, we sequentially connect $T$ processing blocks with identical architecture, but independent learnable parameters.
To restrict the search space we do not include global graph features and do not allow edge updates between blocks. 
This resembles a conventional message-passing architecture.
In the third phase, we have a single readout block, which aggregates node features into graph-level features to make the crystal property prediction.

Figure~\ref{fig:archgn} only specifies the high-level architecture of the GNN and not the concrete instantiations of update functions $\phi_E, \phi_V, \phi_G$ and aggregation functions $\rho_{E \to V}, \rho_{V \to G}$.
To narrow down suitable concrete implementations, we conducted a two-part hyperparameter search on the \texttt{log\_gvrh} MatBench dataset.
First, we searched for categorical hyperparameters, in a greedy stepwise search.
In the second step, we used the Optuna hyperparameter optimization framework~\citep{optuna_2019} and its implementation of the Tree-structured Parzen Estimator (TPE) to find ordinal hyperparameters.\footnote{Ordinal hyperparameters: depth of GNN ($T$), depth of MLPs $\text{MLP}_E, \text{MLP}_V, \text{MLP}_G$, dimensionality of features}

We assume independence between categorical hyperparameters and optimize for each parameter individually, to keep runtimes within limits.
We initially instantiated all update functions $\phi_E, \phi_V, \phi_G$ with $\text{MLP}$s of depth 3.
Extending $\phi_V$ with residual 
$\phi_V(\vb{x}_v, \hat{\vb{x}}_v, \vb{x}_G) = \vb{x}_v + \text{MLP}_V(\hat{\vb{x}}_v)$
and gated node updates
$\phi_V(\vb{x}_v, \hat{\vb{x}}_v, \vb{x}_G) = \text{GRU}(\vb{x}_v, \text{MLP}_V(\hat{\vb{x}}_v))$
both improve performance.
For aggregation function $\rho_V, \rho_G$ we tried mean, sum, and attention-based functions.
Batch and graph normalization~\citep{cai2021graphnorm} increased training time significantly without considerable benefits for predictive performance.
Prediction accuracy further increases when including atom features (atomic mass, radius, electronegativity, ionization, and oxidation states) in addition to the atomic number.
The influence of the choice of edge selection methods shown in Figure~\ref{fig:edgeselection-gn} is discussed in detail in Section~\ref{sec:connectivity}.

For the $k$NN-based edge selection preprocessing with $k=24$, we searched for ordinal hyperparameters with a hyperparameter optimization.

\paragraph{Embedding Block}
\begin{description}
	\vspace{-1em}
	\item $\phi_E$: Gauss basis expansion of edge distances with \(32\) Gaussians evenly spread on the \([0,8]\) \AA\ interval. As only distance information is used, this embedding is E(3)-invariant.
		The initial representation is projected into a $128$-dimensional embedding space with a single perceptron layer.
	\item $\phi_V$: Embedding of atom features (atomic number, atomic mass, atomic radius, electronegativity, ionization states, and oxidation states) into a $128$-dimensional space.
\end{description}

\paragraph{Processing Blocks}
Five processing blocks are concatenated with identical configurations ($T = 5$).
\begin{description}
	\vspace{-1em}
	\item $\phi_E(\vb{x}_{e_{ij}}, \vb{x}_{v_i}, \vb{x}_{v_j}) = \text{MLP}_E(\vb{x}_{e_{ij}} || \vb{x}_{v_i} || \vb{x}_{v_j}) = \vb{x}'_{e_{ij}}$
		The edge update function, which constructs the message between two nodes, is a five-layer MLP and takes the concatenation of edge, receiver, and sender node features.
	\item $\rho_{E \to V} (\{\vb{x}'_{e_{ij}} | j = k\}) = \underset{j = k}{\text{sum}}(\vb{x}'_{e_{ij}}) = \hat{\vb{x}}_{v_k}$
		For each node the incoming messages are sum-aggregated.
	\item $\phi_V(\vb{x}_v, \hat{\vb{x}}_v) = \vb{x}_v + \text{MLP}_V(\hat{\vb{x}}_v)$ 
		A residual node update function transforms the aggregated messages with a single-layer perceptron.
\end{description}

\paragraph{Readout Block}
\begin{description}
	\vspace{-1em}
	\item $\phi_V(\vb{x}_v) = \vb{x}_v$
		The readout block does not update node features. Its node update function is the identity function.
	\item $\rho_{V \to G}(\{\vb{x}'_v | v \in V \}) = \underset{v \in V}{\text{mean}}(\vb{x}'_v) = \hat{\vb{x}}_G$
		The mean of the node features is aggregated to compute the graph-level prediction.
	\item $\phi_G(\hat{\vb{x}}_G)  = \text{MLP}_G(\hat{\vb{x}}_G) = \vb{x}'_G$
		A single-layer MLP with a linear activation function creates the final prediction for the crystal property.
\end{description}

We used the same dimensionality (128) for all hidden representations of edges, nodes, and graphs.
Unless mentioned otherwise we use the commonly used swish activation function in MLPs.
The GNN is trained with an Adam optimizer with a linear learning rate scheduler for 800 epochs. 

Overall, coGN is comparably simple, as it only contains MLPs as update functions, mean or sum aggregation functions and no sophisticated message-passing scheme, such as edge-gated or attention-based message passing used in CGCNN, ALIGNN or GeoCGNN.
In this regard, coGN also does not incorporate any domain-specific design decisions, which are not justified by the hyperparameter optimization.

When compared with previous models, we find that along with a connectivity optimization, discussed in section~\ref{sec:connectivity}, a much deeper edge update network $\phi_E$, in our case five layers, yields better results. We attribute this observation to the increased complexity of edge information in a periodic multi-graph that in practice exhibits a large number of edges per node.

}

\paragraph{Nested Graph Network Exploration}{

Based on the results of the first part of the hyperparameter search for parameters with categorical values, we augment the processing blocks with nested GN blocks shown in Figure~\ref{fig:ngn-variants}.

\begin{figure*}[!h]
	\centering
	\begin{subfigure}[b]{0.3\textwidth}
		\centering
		\def\svgwidth{\textwidth}
		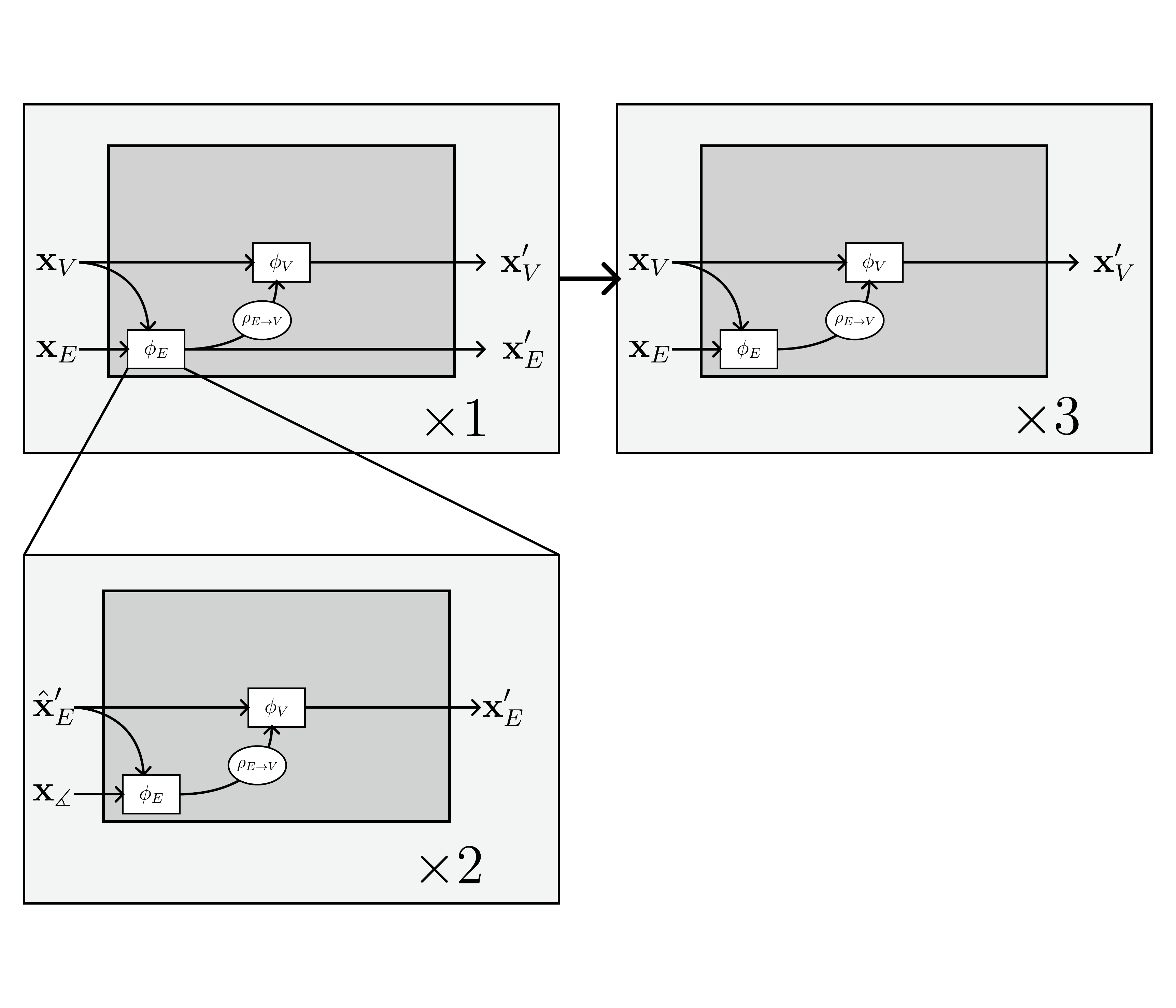
		\caption{NGN variant 1.}
		\label{fig:ngn-variant1}
	\end{subfigure}
	\begin{subfigure}[b]{0.3\textwidth}
		\centering
		\def\svgwidth{\textwidth}
		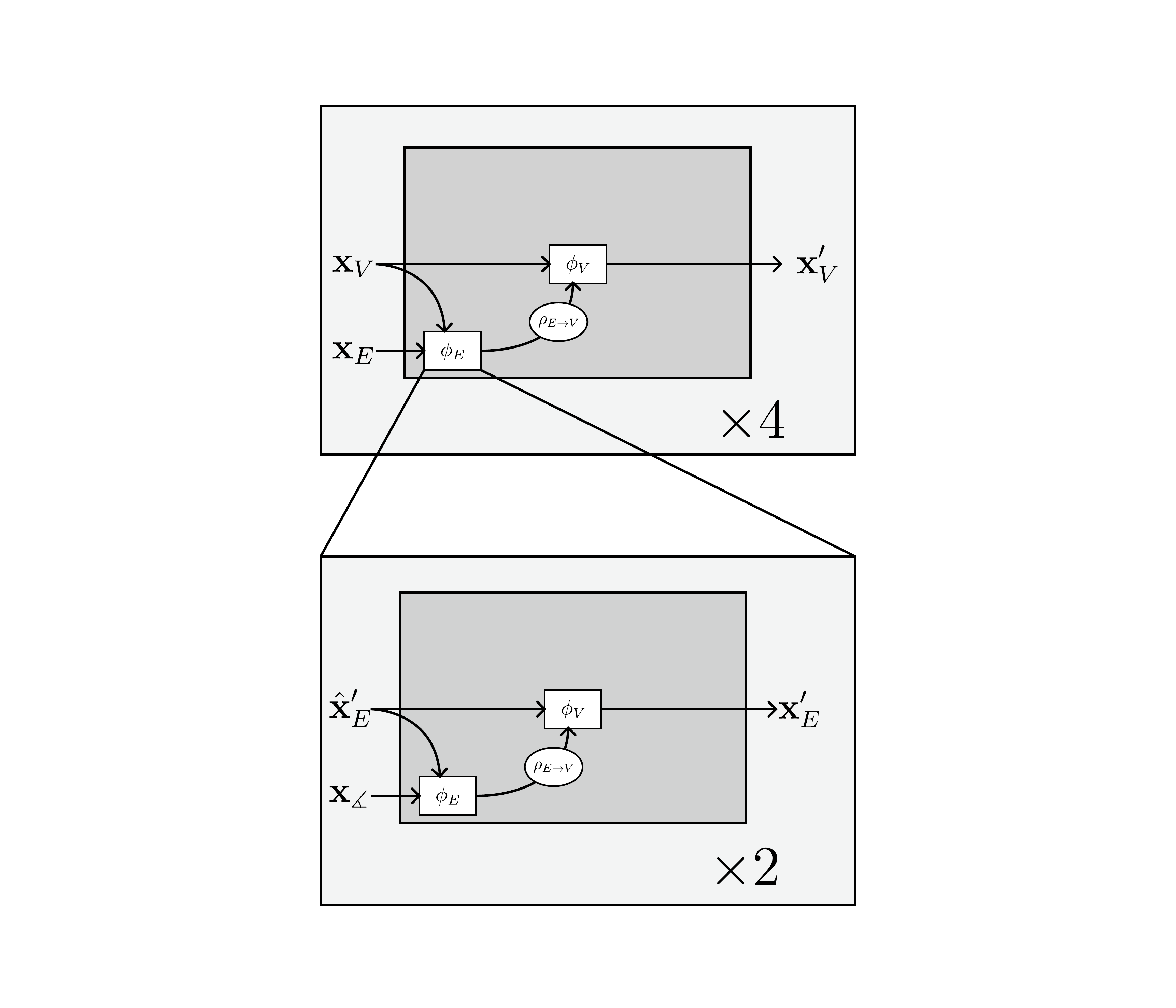
		\caption{NGN variant 2.}
		\label{fig:ngn-variant2}
	\end{subfigure}
	\begin{subfigure}[b]{0.3\textwidth}
		\centering
		\def\svgwidth{\textwidth}
		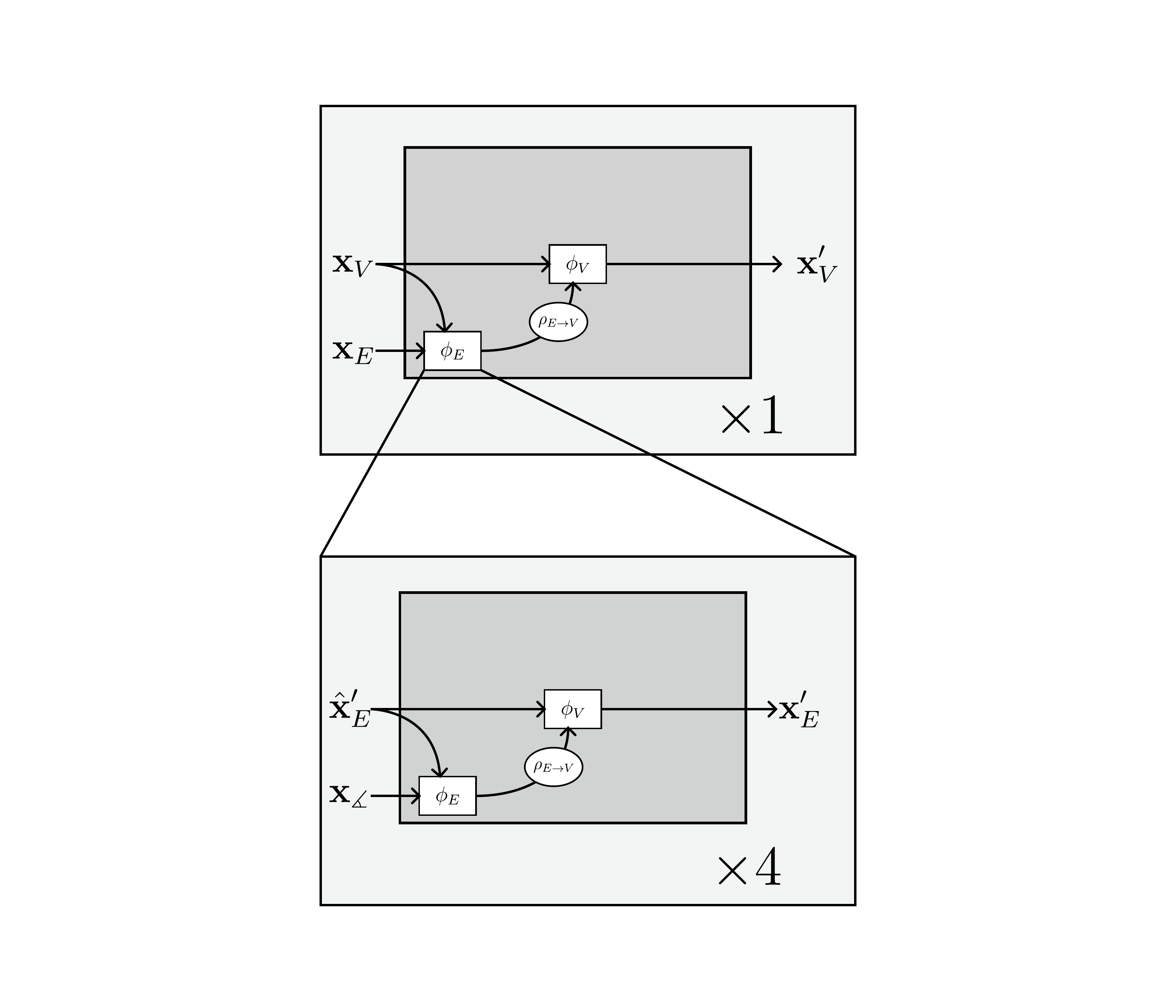
		\caption{NGN variant 3.}
		\label{fig:ngn-variant3}
	\end{subfigure}
	\caption{Nested graph network variants.}
	\label{fig:ngn-variants}
\end{figure*}

The first variant (Figure~\ref{fig:ngn-variant1}) has a single Nested GN block at the beginning, which updates node and edge features, followed by non-nested GN blocks.
The reasoning behind this architecture is that the Nested GN block might be able to encode geometrical constellations of edge angles into edge features.
The second variant (Figure~\ref{fig:ngn-variant2}) is a generalized version of
ALIGNN-d~\citep{hsu2021efficient} and DimeNet(\texttt{++})~\citep{klicpera2020directional,klicpera2020fast}. Each GN block on the graph level (\(G\)) has two GN blocks on the line graph level
(\(L(G)\)). Only node features are updated between GN blocks.
The third variant (Figure~\ref{fig:ngn-variant3}) moves most of the
computation to the line graph level (\(L(G)\)). It consists of only one
GN block on the graph level with 4 consecutive nested GN blocks ($T^{L(G)} = 4$).

E(3)-invariant angle features between two edges are encoded with a 16-dimensional Gauss basis expansion on the \([0,\pi]\) rad interval and attached to line graph edges.

\begin{table}[h]
	\centering
	\caption{Results on the \texttt{log\_gvrh} dataset with NGN variants. 
 }
	\begin{tabular}{lll}
		\toprule
		Nested GN & Line Graph & MAE (\texttt{log\_gvrh}) \\
		\midrule
		\multirow{2}{4em}{Variant 1}   & $\measuredangle e_{ij}e_{jk}$          &  $0.0809\pm0.0022$\\
									   & $\measuredangle e_{ij}e_{kj}$          &  $0.0787\pm0.0019$\\
									   \midrule
		\multirow{2}{4em}{Variant 2}   & $\measuredangle e_{ij}e_{jk}$          &  $0.0801\pm0.0020$\\
									   & $\measuredangle e_{ij}e_{kj}$          &  $0.0783\pm0.0033$\\
									   \midrule
		\multirow{2}{4em}{Variant 3}   & $\measuredangle e_{ij}e_{jk}$          &  $0.0805\pm0.0015$\\
									   & $\measuredangle e_{ij}e_{kj}$          &  $0.0799\pm0.0016$\\
									   \bottomrule
	\end{tabular}
	\label{tab:ngn-variants}
\end{table}

Training on preprocessed crystals with $k$NN edge selection and $k=24$ and both line graph variants from Section~\ref{sec:linegraph}, we obtain the results displayed in Table~\ref{tab:ngn-variants}.
The results were obtained before the final ordinal hyperparameter optimization, which explains the discrepancy with Table~\ref{tab:MatBench_results}. For comparison, the MAE for the corresponding non-nested GN is $0.0788$.

Despite the theoretically greater expressiveness of NGNs, we do not achieve substantially better prediction results.
The proposed line graph variant with angles between edges with the same target node ($\measuredangle e_{ij}e_{kj}$) leads to a small but consistent improvement across all line graph variants.

We optimized the ordinal hyperparameters of the DimeNet-like architecture (Variant 2) with TPE and reached a
MAE of $0.0705$ on the \texttt{log\_gvrh} MatBench dataset which yields better performance than the current leader on MatBench~\citep{choudhary2021atomistic}, even though the graph preprocessing, i.e.\ the connectivity was not optimized yet.
Hyperparameter optimization of plain GNs yields a similar error of $0.0693$, making them comparable to Nested GNs.

}

\subsection{Crystal Graph Connectivity}\label{sec:connectivity}

For GNNs in particular, there is a strong interdependency between input representation and model, because the topology of the input graph also affects the computational graph.
The interdependency between preprocessing of crystals and model architecture also occurs with respect to crystal property predictions and should therefore be considered.

\begin{figure}[h]
    \centering
	\def\svgwidth{.55\textwidth}
	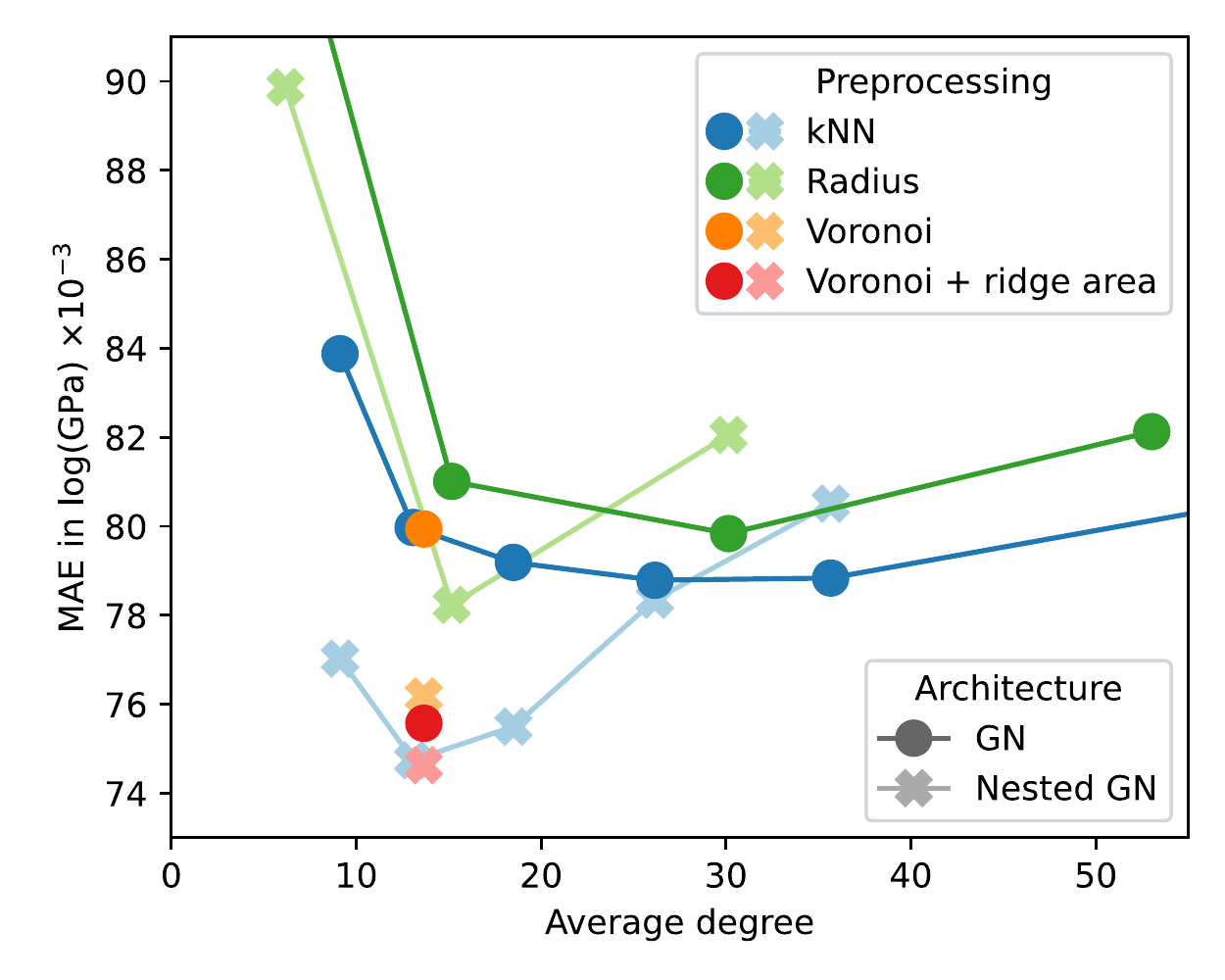
	\caption{MAE for GNs, NGNs, and different edge selection methods on the \texttt{log\_gvrh} dataset.}
    \label{fig:edgeselection-gn}
\end{figure}

Figure~\ref{fig:edgeselection-gn} shows the effect of different preprocessing methods for a non-nested GN. Again, results were obtained before the final ordinal hyperparameter optimization, which explains the discrepancy with Table~\ref{tab:MatBench_results}.
The MAE is plotted as a function of the average degree over the resulting graphs of the \texttt{log\_gvrh} dataset for different edge selection methods.
The accuracy of GNN models tends to fall for increasing graph connectivity up to an average degree of about 30.
For a higher average degree the accuracy either gets worse or quickly converges depending on the dataset.
Interestingly, for the $k$NN edge selection the minimum corresponds to $k = 24$, whereas other works use a value of $12$ \citep{xie2018crystal, choudhary2021atomistic}.
The Voronoi-based edge selection results in graphs with an average degree of around $12$.
Adding the area of the Voronoi cell ridge to each edge as an additional edge feature can improve the predictive power of the GNN.
We found, however, that this effect is much less pronounced after optimization of ordinal hyperparameters, making the $k=24$ nearest-neighbor method the preferred choice for edge selection in our experiments.

The observations for non-nested GNs do not apply to NGNs, when comparing their behavior in Figure~\ref{fig:edgeselection-gn}.
For NGNs, the minimum test error occurs at a lower average degree of around $12$ edges per node and is followed by a steeper increase for higher connectivity.
Higher-order nesting could potentially lead to further improvement at yet lower connectivity of the base graph, which should be systematically explored in the future.

\begin{figure}[h]
	\centering
	\def\svgwidth{.5\textwidth}
	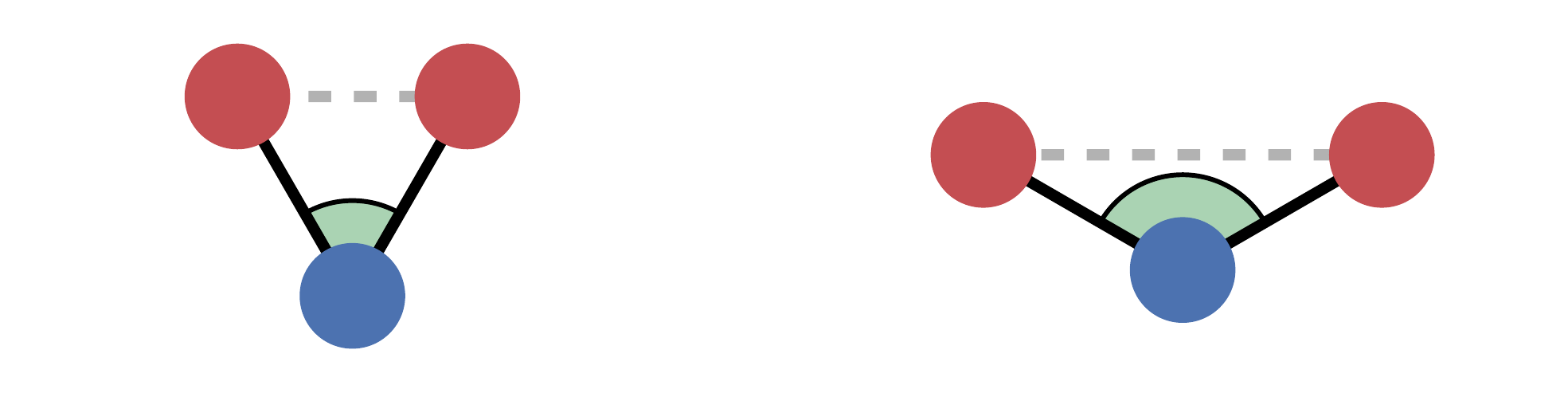
	\caption{Two different geometric graphs which are undistinguishable from distance edge features, but become distinguishable with angle information between edges or by adding the dashed edge to the graphs.}
	\label{fig:expressiveness}
\end{figure}

At the same time, our observations raise the issue of a trade-off between nesting and graph connectivity.
This trade-off can also be explained from an analytical point of view.
The reason for including angle information with NGNs is shown in the example in Figure~\ref{fig:expressiveness}.
The two geometric graphs are not distinguishable for GNNs from relative distance information alone.
Incorporating angles between edges into the GNN architecture increases expressiveness and allows for the distinction of the graphs.
Yet, similar enhancements of expressiveness can also be achieved by increasing graph connectivity.
In the example, adding the single dashed edge between two red nodes makes the graphs distinguishable for GNNs, without any angle information.

From our experiments, we can not conclude that NGNs offer a systematic advantage over simple GNs when used on graphs with high connectivity. 
Inspired by the results of Figure~\ref{fig:edgeselection-gn}, we optimized an NGN for less connected Voronoi (+ ridge area) preprocessed graphs and achieved the results displayed in Table~\ref{tab:MatBench_results}.
Although NGNs showed the best performance on the specific dataset they have been (hyperparameter) optimized on, they can not maintain their advantage consistently on all other datasets without re-optimizing.
Unfortunately, NGNs require the construction of line graphs and tend to have significantly more trainable parameters.
Consequently, training is more than three times as computationally expensive as for non-nested GNs. 

Finally, we were able to demonstrate the effectiveness of densely connected crystal graphs with the coGN model, by surpassing current state-of-the-art models on most of the MatBench datasets as shown in Table~\ref{tab:MatBench_results}.
The results also support that hyperparameters, which originate from the optimization on the \texttt{log\_gvrh} dataset, generalize to other datasets and tasks.
Further results and statistics can be found in the Appendix~\ref{app:oc22}.

\begin{table*}[h]
	\centering
	\caption{Comparison with results of state-of-the-art model on MatBench structure datasets and splits~\citep{dunn_benchmarking_2020} (status 2023-08-08). Ordered by descending cardinality, these are \texttt{e\_form} (meV/atom), \texttt{is\_metal} (AUC(ROC)), \texttt{gap} (meV), \texttt{perovskites} (meV/unit cell), \texttt{log\_kvrh} (log10(GPa)), \texttt{log\_gvrh} (log10(GPa)), \texttt{dielectric} (unitless), \texttt{phonons} (1/cm) and \texttt{jdft2d} (meV/atom). Current benchmark holders on MatBench, namely, ALIGNN~\citep{choudhary2021atomistic}, MODNet~\citep{de_breuck_materials_2021} and CGCNN~\citep{xie2018crystal} are listed. Additionally, recent models M3GNet~\citep{chen2022universal} and Matformer~\citep{yan2022periodic} are added, which have been published during the preparation of this work. Since Matformer was not trained on the official benchmark, we retrained the original model. The best results are indicated in bold font, while other results within one standard deviation are underlined.
    $^*$For \texttt{is\_metal} there is a discrepancy between the submissions for AUC computation (which is why the classification metric is likely to change in future version).
	}
    \resizebox{\textwidth}{!}{
	\begin{tabular}{c||c|c|c|c|c|c|c}
		\toprule
		Dataset & coGN (ours) &  coNGN (ours) & ALIGNN &  MODNet & CGCNN  &  M3GNet & Matformer\\
		\midrule
\texttt{e\_form} \textdownarrow      & \textbf{17.0 ± 0.3} &         17.8 ± 0.4  & 21.5 ± 0.5       &         44.8 ± 3.9        &         33.7 ± 0.6       & 19.5 ± 0.2    &  21.232 ± 0.302  \\
\texttt{is\_metal$^*$} \textuparrow   &         0.9124 ± 0.0023  &         0.9089 ± 0.0019                & 0.9128 ± 0.0015  &         0.9038 ± 0.0106   & \textbf{0.9520 ± 0.0074}  & 0.958±0.001   & 0.906 ± 0.002   \\
\texttt{gap} \textdownarrow        & \textbf{155.9 ± 1.7} &         169.7 ± 3.5  & 186.1 ± 3.0      &         219.9 ± 5.9       &         297.2 ± 3.5      & 183 ± 5       & 187.825 ± 3.817 \\
\texttt{perovskites} \textdownarrow & \textbf{26.9 ± 0.8} &         29.0 ± 1.1  & 28.8 ± 0.9       &         90.8 ± 2.8        &         45.2 ± 0.7       & 33 ± 1.0 & 31.514 ± 0.71 \\
\texttt{log\_kvrh} \textdownarrow  &         0.0535 ± 0.0028  & \textbf{0.0491 ± 0.0026} & 0.0568 ± 0.0028  &         0.0548 ± 0.0025   &         0.0712 ± 0.0028  & 0.058±0.003   & 0.063 ± 0.0027\\
\texttt{log\_gvrh} \textdownarrow  &        0.0689 ± 0.0009  & \textbf{0.0670 ± 0.0006} & 0.0715 ± 0.0006  &         0.0731 ± 0.0007   &         0.0895 ± 0.0016  & 0.086±0.002   & 0.077 ± 0.0016 \\
\texttt{dielectric} \textdownarrow &         \underline{0.3088 ± 0.0859}  &         \underline{0.3142 ± 0.0740}  & 0.3449 ± 0.0871  & \textbf{0.2711 ± 0.0714}  &         0.5988 ± 0.0833  & 0.312±0.063   & 0.634 ± 0.131 \\
\texttt{phonons} \textdownarrow    &         \underline{29.712 ± 1.997}  & \textbf{28.887 ± 3.284} & \underline{29.539 ± 2.115} &         34.2751 ± 2.0781  &         57.7635 ± 12.311 & 34.1 ± 4.5    & 42.526 ± 11.886\\
\texttt{jdft2d} \textdownarrow     &         \underline{37.165 ± 13.683}  &         \underline{36.170 ± 11.597}  & 43.424 ± 8.949 & \textbf{33.192 ± 7.343} &         49.244 ± 11.587 & 50.1 ± 11.9    & 42.827 ± 12.281\\
		\bottomrule
	\end{tabular}}
	\label{tab:MatBench_results}
\end{table*}

\subsection{Asymmetric Unit Graphs}

In Section~\ref{sec:crystal-symmetries}, we discussed different graph representations for crystals and found that asymmetric unit graphs are smaller than unit cell graphs and yet lead to identical predictions of GNNs with E(3)-invariant layers.
The exact reduction factor for asymmetric unit graphs depends on the symmetries each specific crystal exhibits.
Since the maximal space group order for crystals is 48, we can specify theoretical lower and upper bounds for the number of nodes $n_\text{asu}$ in asymmetric unit graphs in relation to unit cell graphs:
\begin{equation*}
\protect\hypertarget{eq:node-rel}{}{n_\text{unit} \geq n_\text{asu} \geq \frac{1}{48} \cdot n_\text{unit}}\label{eq:node-rel}
\end{equation*}
We found that for the MatBench datasets, the empirical average reduction factor for the number of nodes ($n$), edges ($m = n^{L(G)}$) as well as line graph edges ($m^{L(G)}$) is approximately 2.1 \footnote{The \texttt{perovskites} dataset is a special outlier, where asymmetric graphs are on average not significantly smaller than unit cell graphs.}:
\begin{equation*}
\frac{n_\text{unit}}{n_\text{asu}} \approx \frac{m_\text{unit}}{m_\text{asu}} \approx \frac{m^{L(G)}_\text{unit}}{m^{L(G)}_\text{asu}} \approx 2.1
\end{equation*}
Approximately the same factor can be observed for the GPU memory footprint during training.
Due to parallelization and batching the acceleration of the training runtimes is somewhat smaller and depends on the selected batch size.
For batch sizes of 32, 64, and 128, for example, we get a speedup of $1.2$, $1.3$, and $1.8$, respectively.

\section{Conclusion}

This paper discusses fundamental aspects of crystal property prediction with GNNs: the incorporation of symmetries in GNN models, the interdependence of input graph generation from crystal structures (i.e.\ preprocessing) with the systematic exploration of a suitable GNN architecture, as well as the generalization of model architectures as nested graph networks.
We conclude that these aspects cannot be considered separately.

Our contribution to the first aspect includes the proposal of the \textit{asymmetric unit graph} representation, which decreases training time and memory consumption without effects on predictive performance for E(3)-invariant GNNs.
To use asymmetric unit graphs with equivariant GNNs some adaptations to message passing and readout operations are required, which will be addressed in future work.
We furthermore compared different edge selection methods and discovered that graphs with higher connectivity (compared to previous works) can yield better performance.
From this insight, we construct the coGN, which achieves state-of-the-art results on the MatBench benchmark.

To explore the space of GNN architectures, we introduced the \textit{nested graph network} (NGN) framework, which subsumes state-of-the-art GNN models which incorporate angle information in line graph form.
We observed that a variation in the line graph definition for angles is likely to improve predictions by a small amount.
Although we could not find an architecture for nested graph networks that substantially outperforms graph networks without nesting, we still encourage further research in this direction, specifically into the scaling laws of accuracy as a function of data set size and complexity as well as the depth of line-graph nesting.
Future work might overcome the mentioned trade-off between high connectivity and nesting and potentially explore specific nested graph network architectures, which we did not consider due to hyperparameter space restrictions.

\section*{Acknowledgement}

The authors acknowledge support by the state of Baden-Württemberg through bwHPC.

\bibliographystyle{unsrtnat}
\bibliography{references}  






\newpage
\section{Appendix}

\FloatBarrier
\subsection{Crystal Preprocessing: Edge Selection}\label{app:edge-selection}

\begin{figure}[h]
\hypertarget{fig:preprocessor-distributions}{%
\centering
\def\svgwidth{.95\textwidth}
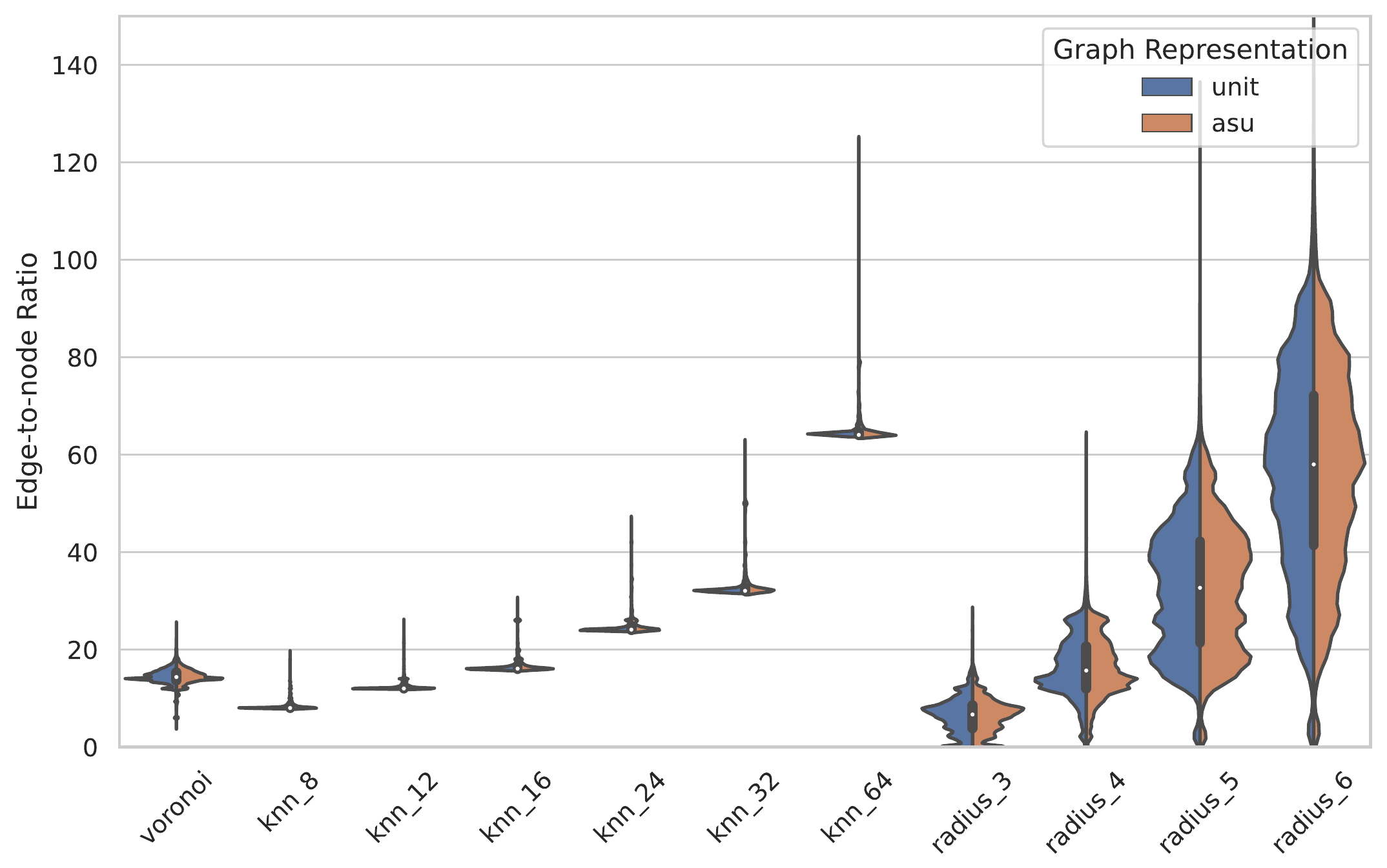
\caption{Distributions of edge-to-node ratios per graph (average degree) in the
\texttt{mp\_e\_form} dataset for different preprocessing
methods. Due to the
symmetrical and regular structure of crystals, it often happens that
there are many neighbors with equal distances. Therefore our
\(k\)NN-based edge selection implementation, also allows to toggle
whether edges with a \(\epsilon\)-distance difference as the \(k\)-th
nearest neighbor are also added as connections. We included an
\(\epsilon = 10^{-9}\) \AA threshold to make up for numerical inaccuracies.}\label{fig:preprocessor-distributions}
}
\end{figure}

\begin{figure}[h]
\hypertarget{fig:distance-distribution}{%
\centering
\def\svgwidth{.95\textwidth}
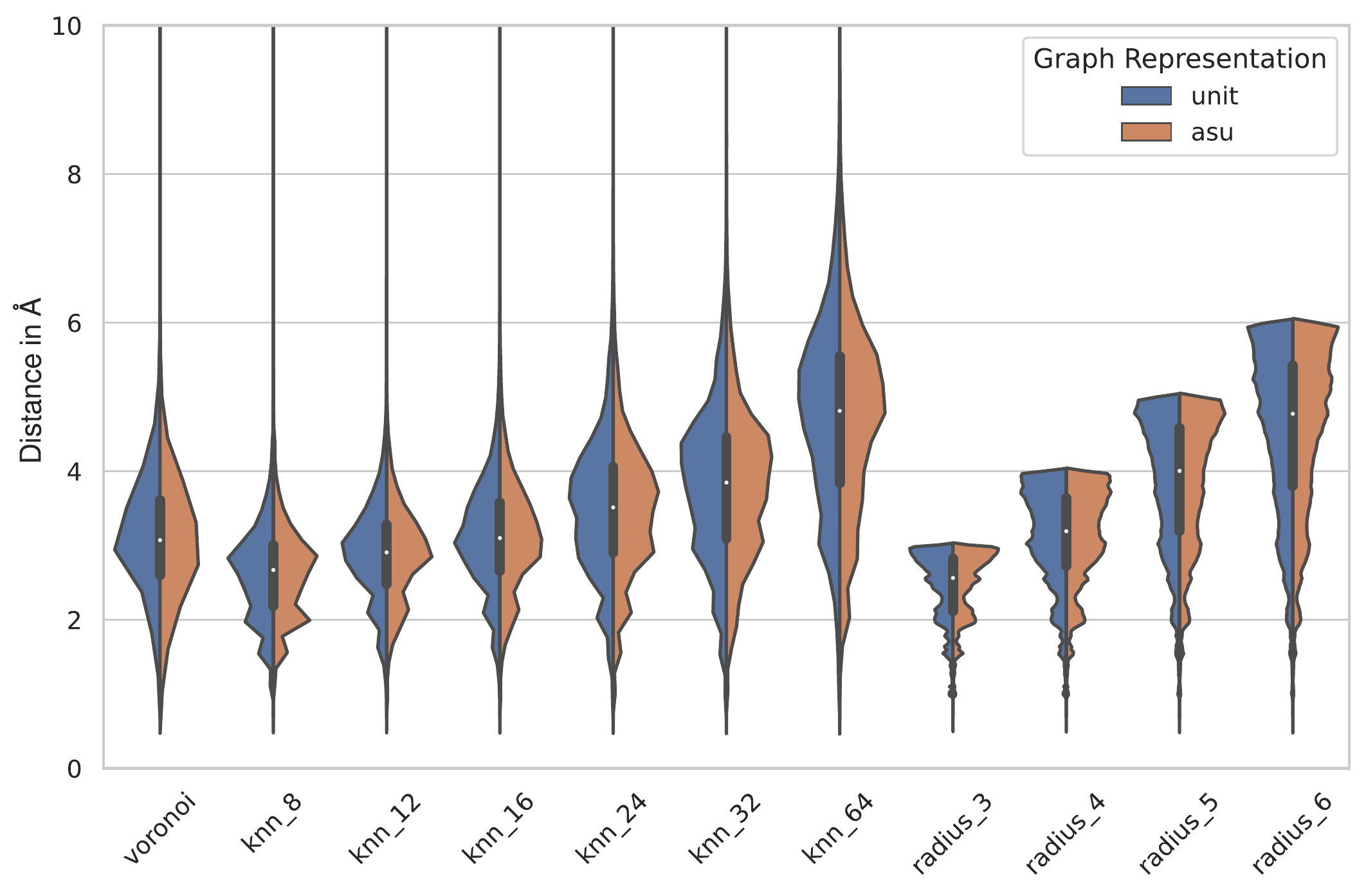
\caption{Distribution of edge distances in the
\texttt{mp\_e\_form} dataset for different edge selection methods.}\label{fig:distance-distribution}
}
\end{figure}

\FloatBarrier
\subsection{Crystal Preprocessing: Symmetries}\label{app:symmetries}

\begin{figure}[h]
\hypertarget{fig:avg-num-nodes}{%
\centering
\def\svgwidth{.95\textwidth}
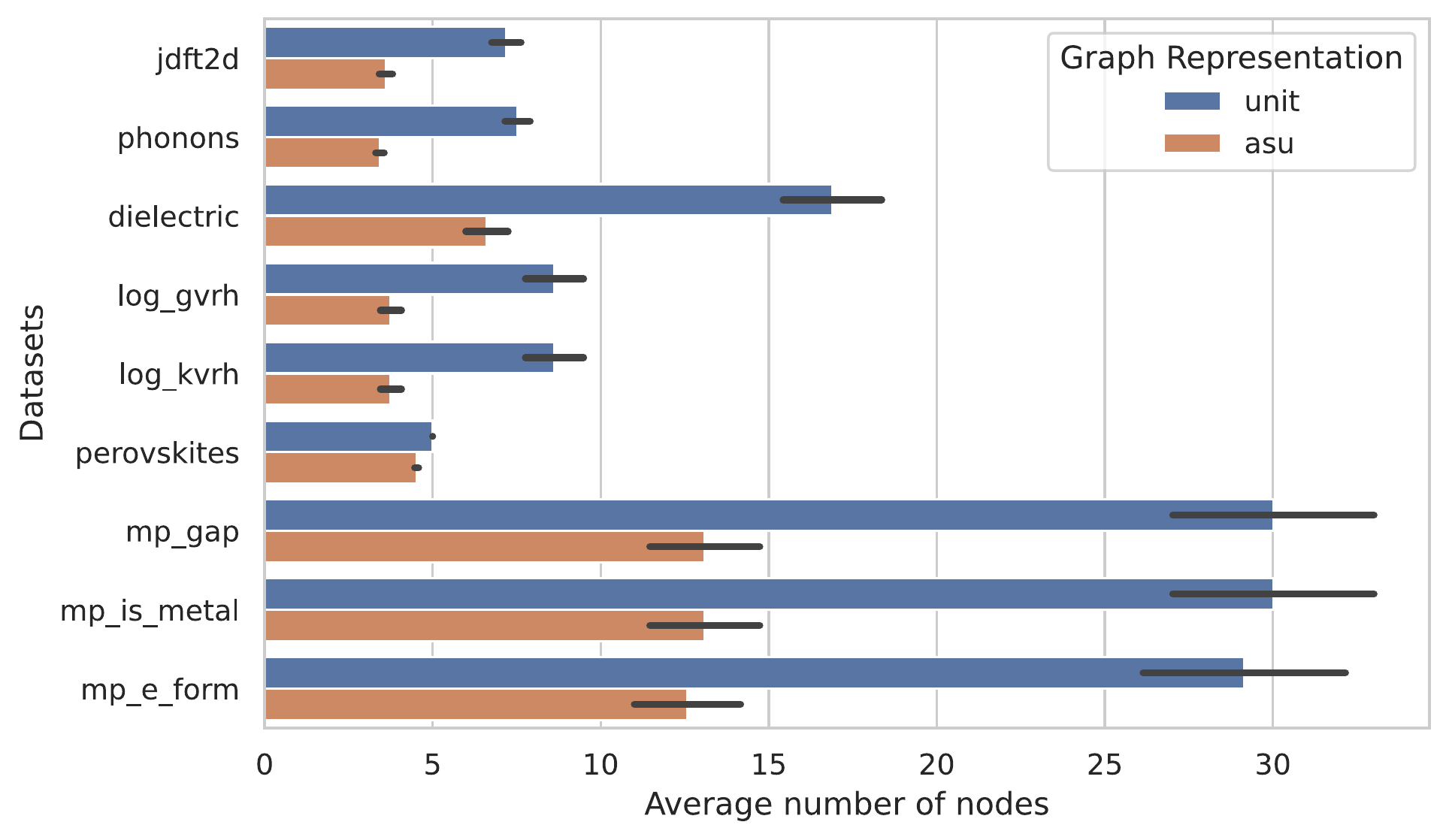
\caption{Average number of nodes in each dataset for unit cell graphs
(\textcolor[HTML]{5875a4}{unit}) and asymmetric unit graphs (\textcolor[HTML]{cc8963}{asu}).}\label{fig:avg-num-nodes}
}
\end{figure}

\begin{figure}
\hypertarget{fig:hist-num-nodes}{%
\centering
\def\svgwidth{.95\textwidth}
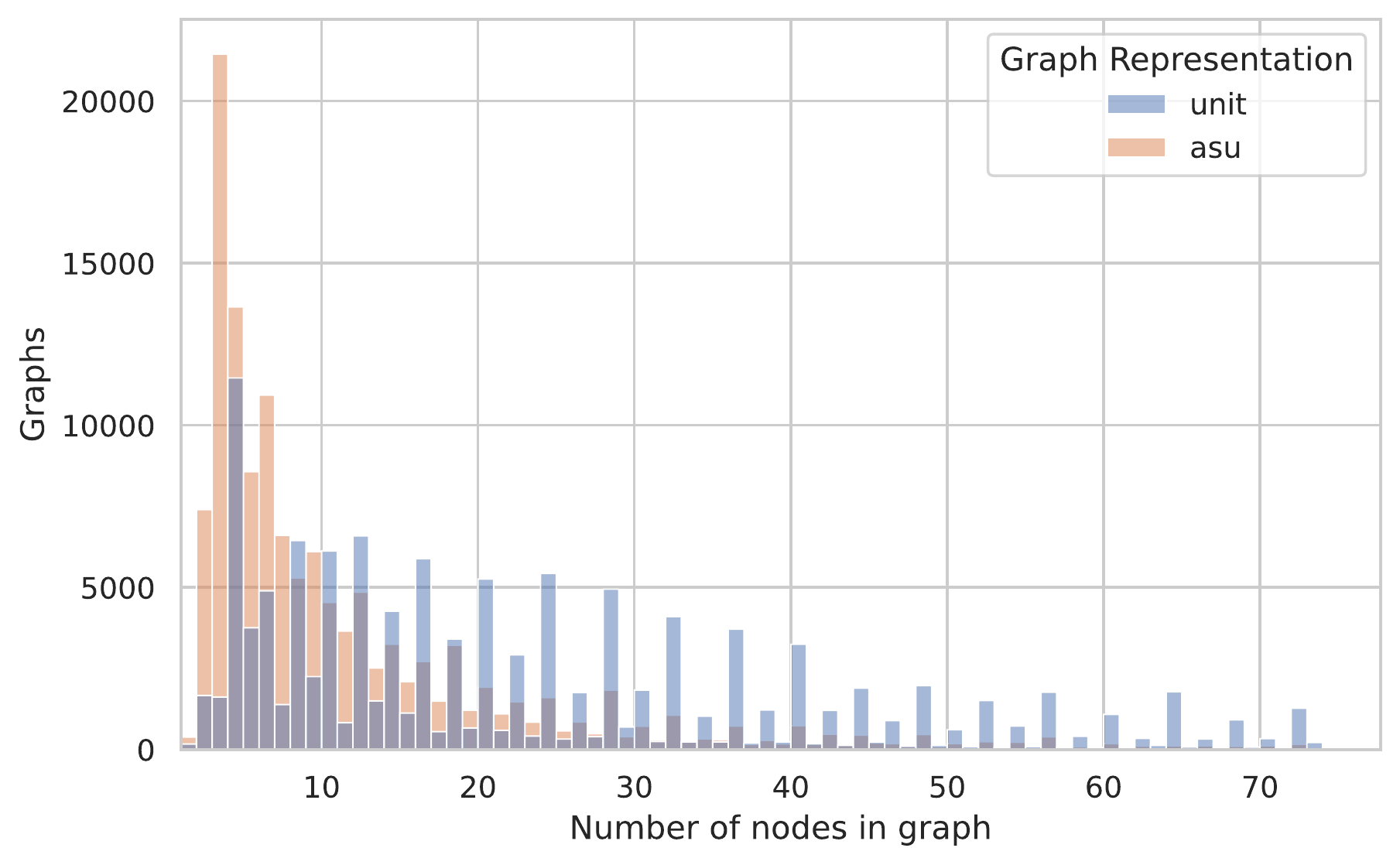
\caption{Histogram of number of nodes in \texttt{mp\_e\_form} dataset
for unit cell graphs (\textcolor[HTML]{5875a4}{unit}) and asymmetric unit graphs
(\textcolor[HTML]{cc8963}{asu}). The distribution for asymmetric unit graphs is skewed towards lower number of nodes. Clearly pronounced periodic peaks at even numbers for the unit cell graph distribution can be explained with symmetric atom pairs.}\label{fig:hist-num-nodes}
}
\end{figure}

\FloatBarrier
\subsection{Nested Graph Network Algorithm}\label{app:ngnalg}

\begin{algorithm}
\caption{Graph network algorithm adapted from  \citep{battaglia2018relational} with additional \textcolor{blue}{blue parts} that indicate the modifications for the \textcolor{blue}{nested graph networks}. GN blocks are parameterized by parameters $\theta$ and receive edge features $\vb{X}_E$, node features $\vb{X}_V$, graph level features $\vb{x}_G$ as well as the graph topology $G$ as input.}\label{alg:gn}
\DontPrintSemicolon 
\SetKwFunction{FMain}{\textcolor{blue}{N}GN-Block}
  \SetKwProg{Fn}{Function}{:}{}
  \SetKwProg{Pn}{Function}{:}{\KwRet $\vb{X}'_E,\vb{X}'_V,\vb{x}'_G,G$}
  \Pn{\FMain{$\vb{X}_E, \vb{X}_V, \vb{x}_G, G; \theta$}}{
        \tcp{$\vb{X}_E = \{x_{e_1}, \ldots, x_{e_m}\}$}
        \tcp{$\vb{X}_V = \{x_{v_1}, \ldots, x_{v_n}\}$}
        
        \For{$e_{ij} \in E$}{
            $\vb{x}'_{e_{ij}} \gets \hat{\phi}_E(\vb{x}_{e_{ij}}, \vb{x}_{v_i}, \vb{x}_{v_j}, \vb{x}_G)$ \tcp*{Edge update}
        }
        \textcolor{blue}{
        $\vb{X}_\measuredangle \gets$ \texttt{get\_line\_graph\_edge\_features}$(\vb{X}_E, \vb{X}_V, \vb{x}_G, G)$ \tcp*{i.e. angles between edges}
        \For{$t \gets 1$ \KwTo $T^{L(G)}$}{
            \_~, $\vb{X}'_E$~, \_~, \_ $\gets$ \texttt{GN-Block}$(\vb{X}_\measuredangle, \vb{X}'_E, \vb{x}_G, L(G);~\theta^t)$ \tcp*{Treat $\vb{X}'_E$ as node features of $L(G)$}
        }}
        \For{$v_k \in V$}{
   	        $\hat{\vb{x}}_{v_k} \gets \rho_{E \to V}(\{\vb{x}'_{e_{ij}} | e_{ij} \in E \land j = k\})$
            \tcp*{Local edge aggregation}
	        $\vb{x}'_{v_k} \gets \phi_V(\vb{x}_{v_k}, \hat{\vb{x}}_{v_k}, \vb{x}_G)$ \tcp*{Node update}
        
        }
        $\hat{\vb{x}}_G \gets \rho_{V \to G}(\{\vb{x}'_v | v \in V \})$ \tcp*{Node aggregation}
        $\tilde{\vb{x}}_G \gets \rho_{E \to G}(\{\vb{x}'_e | e \in E\})$ \tcp*{Global edge aggregation}
        $\vb{x}'_G \gets \phi_G(\vb{x}_G, \hat{\vb{x}}_G, \tilde{\vb{x}}_G)$ \tcp*{Global update}
        \tcp{$\vb{X}'_E = \{x'_{e_1}, \ldots, x'_{e_m}\}$}
        \tcp{$\vb{X}'_V = \{x'_{v_1}, \ldots, x'_{v_n}\}$}
}
\BlankLine
\BlankLine
\tcp{Iterate over sequentially composed \textcolor{blue}{N}GN blocks}
\For{$t \gets 1$ \KwTo $T$}{
            $\vb{X}_E,\vb{X}_V,\vb{x}_G,G \gets$ \texttt{\textcolor{blue}{N}GN-Block}$\left(\vb{X}_E,\vb{X}_V,\vb{x}_G,G ; \theta_t \right)$
}
\;
\end{algorithm}

\FloatBarrier
\subsection{Open Catalyst Project: OC22}\label{app:oc22}

\begin{table*}[h]
	\centering
	\caption{Test error for the \textit{initial structure to relaxed energies} (IS2RES) task of the OC22 challenge~\citep{oc22_dataset} (status 2023-08-08). For comparison, this table includes solely direct OC22-only predictions. Models trained additionally on OC20~\citep{ocp_dataset} and other data sources or indirect predictions using relaxations yield lower errors and better performance. Baseline models are SchNet~\citep{schutt2017schnet}, DimeNet\texttt{++}~\citep{Klicpera2020DimeNet}, PaiNN~\citep{schutt2021equivariant} and GemNet~\citep{klicpera2021gemnet}.
	}
	\begin{tabular}{c||c|c|c}
		\toprule
		Model & MAE (ID) &  MAE (OOD) & Average \\
		\midrule
        coGN (Direct OC22-only, r=5.0) & \underline{\textbf{1.6183}} & \underline{\textbf{2.8058}} &  \underline{\textbf{2.2121}} \\
        coGN (Direct OC22-only, k=32) & 1.6278 & 2.9706 & 2.2992 \\
        GemNet-dT (Direct OC22-only)  & 1.6771 & 3.0837 & 2.3804 \\
        PaiNN (Direct OC22-only)  &  1.716  &  3.6835 & 2.6997 \\
        DimeNet\texttt{++} (Direct OC22-only) & 1.96 &  3.5186 & 2.7393 \\
        SchNet (Direct OC22-only) & 2.0012  &  4.8468 & 3.424 \\
		\bottomrule
	\end{tabular}
	\label{tab:OC22}
\end{table*}

\begin{table*}[h]
	\centering
	\caption{Test error for multiple tasks of the JARVIS benchmark~\citep{choudhary_joint_2020}. Values are copied from the JARVIS leaderboard (status 2023-08-08). Models for comparison are DimeNet\texttt{++} (DN\texttt{++})~\citep{Klicpera2020DimeNet} , ALIGNN~\citep{choudhary2021atomistic}, CGCNN~\citep{xie2018crystal}, MatMiner (MM)~\cite{li_critical_2023, matminer_toolkit} and CFID~\citep{PhysRevMaterials.2.083801}.
	}
	\begin{tabular}{c||c|c|c|c|c|c|c}
		\toprule
		Task & coGN &  coNGN & DN\texttt{++} & ALIGNN & CGCNN & MM & CFID\\
		\midrule
         dft\_3d\_mepsz & 24.1081 & \underline{\textbf{22.842}} & 30.3644 & 23.7313 &  36.0538 & 24.6651 &  29.3445 \\
         dft\_3d\_exfoliation\_energy & 47.6979 & 46.272 & 46.1517 & 52.7033 & 52.7033 & \underline{\textbf{40.887}} &  62.1169 \\
         dft\_3d\_shear\_modulus\_gv & 8.6612 & \underline{\textbf{8.4881}} & 26.0817 & 9.476 & 16.0459 & 10.5415 &  11.9164\\
         dft\_3d\_spillage & 0.3609 & \underline{\textbf{0.3463}} & 0.4137 & 0.351 & 0.3965 & 0.3592 &  0.3867\\
         dft\_3d\_optb88vdw\_total\_energy & \underline{\textbf{0.0262}} & 0.0273 & 0.051 & 0.0367 & 0.0815 & 0.0936 &  0.2436 \\
         dft\_3d\_mepsx & 24.2289 & \underline{\textbf{23.3801}} & 31.9568 & 24.0458 & 33.6597 & 25.2932 & 30.261\\
         dft\_3d\_epsz & 19.6192 & \underline{\textbf{17.8104}} & 33.8379 & 19.5678 & 33.6597 & 25.2932 & 24.781\\
         dft\_3d\_dfpt\_piezo\_max\_dij & 15.2235 & \underline{\textbf{13.8868}} &  13.9889 & 20.5705 & 16.0135 & 21.5729 & -\\
         dft\_3d\_mepsy & 24.1891 & \underline{\textbf{23.3299}} & 31.0215 & 23.6482 & 32.4577 & 25.0706 & 30.0578\\
         qe\_tb\_indir\_gap & 0.0474 & - & - & 0.1167 & - & \underline{\textbf{0.0351}} & -\\
         dft\_3d\_kpoint\_length\_unit & 9.5722 & 9.3459 & 11.8875 & 9.5146 & 13.2145 &	\underline{\textbf{9.047}} & 9.7085\\
         dft\_3d\_n\_powerfact & 452.235 & 456.6118 & 568.8357 & \underline{\textbf{442.2993}} & - & 469.6279 & - \\
         dft\_3d\_ph\_heat\_capacity & 6.1125 & 7.8127 & 23.3618 & 9.6064 & - & 	\textbf{5.2757} & -\\
         dft\_3d\_formation\_energy\_peratom & \underline{\textbf{0.0271}} & 0.0291 & 0.0528 & 0.0331 & 0.0625 & 0.0734 & 0.1419\\
         dft\_3d\_epsx & 20.0004 & \underline{\textbf{18.5738}} & 27.2511 & 20.3942 & 31.4744 & 21.2597 & 24.8408\\
         dft\_3d\_optb88vdw\_bandgap & \underline{\textbf{0.1219}} & 0.1267 & 0.2247 & 0.1423 & 0.1908 & 0.1873 & 0.299\\
         qe\_tb\_energy\_per\_atom & \underline{\textbf{0.0636}} &  - & 0.7515 & - & - & 1.5049 & -\\
         dft\_3d\_max\_efg & 20.4417 & 19.5495 & 26.9552 & \underline{\textbf{19.1211}} & 24.6695 & 19.4382 & -\\
         dft\_3d\_epsy & 24.1891 & \underline{\textbf{23.3299}} & 31.0215 & 23.6482 & 32.4577 & 	25.0706 & 30.0578\\
         dft\_3d\_encut & 133.8915 & \underline{\textbf{129.8266}} & 164.315 & 133.7962 & 190.3857 & 138.2769 & 139.4357\\
         dft\_3d\_n\_Seebeck & \underline{\textbf{39.2692}} & 40.0977 & 54.2759 & 40.9214 & 49.3172 & 44.2229 & -\\
         dft\_3d\_ehull & \underline{\textbf{0.0466}} & 0.0485 & 0.3685 & 0.0763 & 0.173 & 0.0601 & -\\
         dft\_3d\_bulk\_modulus\_kv & 8.992 & \underline{\textbf{8.7022}} & 13.3743 & 10.3988 & 19.3028 & 12.7411 & 14.1999\\
         dft\_3d\_avg\_hole\_mass & 0.1372 & 0.1285 & 0.1709  & \underline{\textbf{0.1239}} & - & 0.1529 & -\\
         dft\_3d\_avg\_elec\_mass & 0.0917 & 0.0876 & 0.112 & \underline{\textbf{0.0853}} & - & 0.107 & - \\
         dft\_3d\_mbj\_bandgap & \underline{\textbf{0.264}} & 0.2719 & 0.4764 & 0.3104 & 0.4067 & 0.3392 & 0.5313\\
         dft\_3d\_dfpt\_piezo\_max\_dielectric & 30.2923 & \underline{\textbf{25.5553}} & 30.3358 & 	28.1514 & 32.5589 & 36.6913 & - \\
         dft\_3d\_slme & 4.4507 & \underline{\textbf{4.4428}} & 5.6403 & 4.5207 & 5.6603 &4.9255 & 6.2607\\
         qe\_tb\_f\_enp & \underline{\textbf{0.0956}} & - & - & 	0.1016 & - & 	0.3219 & -\\
         dft\_3d\_magmom\_oszicar & 0.2502 & \underline{\textbf{0.2437}} & 0.3995 & 0.2574 & 0.3543 & 0.3645 & 0.4748\\
         qe\_tb\_final\_energy & \underline{\textbf{1.3185}} & - & -& - & - & 1.4714 & -\\
		\bottomrule
	\end{tabular}
	\label{tab:JARVISleader}
\end{table*}

\end{document}